\documentclass[journal]{IEEEtran}
\usepackage{subcaption}
\usepackage{multicol}
\usepackage{tikz, pgfplots}
\usepackage{multirow}
\usepgfplotslibrary{fillbetween}
\usetikzlibrary{patterns}
\usepackage{csquotes}
\MakeOuterQuote{"}

\usepackage{url}

\usepackage{latexsym}
\usepackage{amssymb}
\usepackage{amsmath}
\usepackage{amsthm}
\usepackage{amsfonts}
\usepackage{verbatim}
\usepackage{listings}
\DeclareSymbolFont{largesymbols}{OMX}{cmex}{m}{n} % However for large sigmas we want the Computer Modern symbol

\usepackage{array}
\usepackage{color}
\usepackage{xspace} % xspace takes care of the \@ after a capitalized word before a period.
\usepackage{multicol}
\usepackage{graphicx}
\usepackage{caption}
\usepackage{fancyhdr}
\usepackage{booktabs}
\usepackage{mathtools}

\usepackage[color={red!100!green!33},colorinlistoftodos]{todonotes} % http://www.tex.ac.uk/tex-archive/macros/latex/contrib/
\presetkeys{todonotes}{inline}{}
\newcommand{\punt}[1]{}

\def\var#1{\ifmmode#1\else$#1$\fi}% usual variable
\def\vari#1{\ifmmode\hbox{\it#1}\else{\it#1}\fi}% variable in italic
\def\vartype#1{\ifmmode\hbox{\tt#1}\else{\tt#1}\fi}% variable type

\newcommand{\ang}[1]        {\ifmmode{\left\langle #1 \right\rangle}
                 \else{$\left\langle${#1}$\right\rangle$\xspace}\fi}
\theoremstyle{definition}
\def\squarebox#1{\hbox to #1{\hfill\vbox to #1{\vfill}}}

\newlength{\savearraycolsep}

\renewcommand{\phi}{\varphi}

\renewcommand{\AA}{\mathcal{A}}

\newcommand{\OO}{\mathcal{O}}

\renewcommand{\SS}{\mathcal{S}}

\newcommand{\R}{\mathbb{R}}

\renewcommand{\eqref}[1]      {Equation~(\ref{eq:#1})}

\usepackage{hyperref}
\IEEEoverridecommandlockouts
\makeatletter
\def\ps@IEEEtitlepagestyle{
  \def\@oddfoot{\mycopyrightnotice}
  \def\@evenfoot{}
}
\def\mycopyrightnotice{
  {\footnotesize
  \begin{minipage}{\textwidth}
  \centering
  \vspace{-15mm}
  \copyright 2021 IEEE. Personal use of this material is permitted.  Permission from IEEE must be obtained for all other uses, in any current or future media, including reprinting/republishing this material for advertising or promotional purposes, creating new collective works, for resale or redistribution to servers or lists, or reuse of any copyrighted component of this work in other works. DOI: \href{https://ieeexplore.ieee.org/document/9489303}{10.1109/TRO.2021.3087314}
  \end{minipage}
  }
}

\newcommand{\addspace}{\vspace{1mm}}
\newcommand{\centertab}[1]{\multicolumn{1}{c|}{\textbf{#1}}}

\newcommand{\todoEugene}[1]{\todo[inline,color=green!30!white]{Eugene: #1}}
\newcommand{\todoCathy}[1]{\todo[inline,color=purple!30!white]{Cathy: #1}}
\newcommand{\todoKanaad}[1]{\todo[inline,color=orange!30!white]{Kanaad: #1}}

\newcommand{\todoAboudy}[1]{\todo[inline,color=blue!30!white]{Aboudy: #1}}

\newcommand{\todoKathy}[1]{\todo[inline,color=teal!30!white]{Kathy: #1}}

\newcommand{\eat}[1]{}
\renewcommand{\todoCathy}[1]{}
\renewcommand{\todoAboudy}[1]{}
\renewcommand{\todoEugene}[1]{}
\newcommand{\newCathy}[1]{#1}
\newcommand{\newCathyTwo}[1]{#1}

\usepackage[font=footnotesize]{caption}

\begin{document}
% \title{A new architecture for benchmarking \\ deep Reinforcement Learning \\ with applications to mixed-autonomy traffic}
% \title{Flow: A new architecture for benchmarking \\ deep Reinforcement Learning \\ with applications to mixed-autonomy traffic}
% \title{Flow: an architecture for benchmarking reinforcement learning for traffic control}
% \title{Flow: \newCathy{De-coupling Scenario and Controller Design for Autonomous Vehicle and Traffic Control}}
% \title{Flow: A Modular Reinforcement Learning Framework for Traffic Flow Smoothing with Autonomous Vehicles}
\title{\newCathy{Flow: A Modular Learning Framework for \\ Mixed Autonomy Traffic}}
% \title{Flow: Architecture and Benchmarking for Reinforcement Learning in Traffic Control}
% \title{Framework for Control and Deep Reinforcement Learning in Traffic}

% \author{
%     \IEEEauthorblockN{
%     Cathy Wu\IEEEauthorrefmark{1}, 
%     Aboudy Kreidieh\IEEEauthorrefmark{2},
%     Kanaad Parvate\IEEEauthorrefmark{1},
%     Eugene Vinitsky\IEEEauthorrefmark{3},
%     Alexandre M Bayen\IEEEauthorrefmark{1}\IEEEauthorrefmark{2}\IEEEauthorrefmark{4}
%     }
%     
%     \IEEEauthorblockA{\IEEEauthorrefmark{1}UC Berkeley, Electrical Engineering and Computer Science}
%     \IEEEauthorblockA{\IEEEauthorrefmark{2}UC Berkeley, Department of Civil and Environmental Engineering}
%     \IEEEauthorblockA{\IEEEauthorrefmark{3}UC Berkeley, Department of Mechanical Engineering}
%     \IEEEauthorblockA{\IEEEauthorrefmark{4}UC Berkeley, Institute for Transportation Studies}
% }

\author{
    Cathy Wu, 
    Abdul Rahman Kreidieh,
    Kanaad Parvate,
    Eugene Vinitsky,
    Alexandre M Bayen% <-this % stops a space

% \thanks{\noindent Corresponding author: Cathy Wu (cathywu@mit.edu, 77 Massachusetts Ave, Cambridge, MA 02139)}
% \thanks{Email addresses: \{aboudy, kanaad, evinitsky, bayen\}@berkeley.edu}
    
    \thanks{This research was supported by the National Science Foundation Graduate Research Fellowship Program. \textit{(Corresponding author: Cathy Wu.)}}
    
    \thanks{Cathy Wu is with the Laboratory for Information and Decision Systems, Massachusetts Institute of Technology, Cambridge, MA 02139 USA, with the Department of Civil and Environmental Engineering, Massachusetts Institute of Technology, Cambridge, MA 02139 USA, and also with the Institute of Data, Systems, and Society, Massachusetts Institute of Technology, Cambridge, MA 02139 USA (e-mail: cathywu@mit.edu).}

    \thanks{Abdul Rahman Kreidieh is with the Department of Civil and Environmental Engineering, University of California, Berkeley, Berkeley, CA 94720 USA (e-mail: aboudy@berkeley.edu).}

    \thanks{Kanaad Parvate is with the Department of Electrical Engineering and Computer Sciences, University of California, Berkeley, Berkeley, CA 94720 USA (e-mail: kanaad@berkeley.edu).}
    
    \thanks{Eugene Vinitsky is with the Department of Mechanical Engineering, University of California, Berkeley, Berkeley, CA 94720 USA (e-mail: evinitsky@berkeley.edu).}

    \thanks{Alexandre M. Bayen is with the Department of Electrical Engineering and Computer Sciences, University of California, Berkeley, Berkeley, CA 94720 USA, and also with the Institute for Transportation Studies, University of California, Berkeley, Berkeley, CA 94720 USA (e-mail: bayen@berkeley.edu).}

%    \IEEEauthorblockA{\IEEEauthorrefmark{1}MIT, Laboratory for Information and Decision Systems; \\ MIT, Department of Civil and Environmental Engineering; MIT, Institute of Data, Systems, \& Society} %
%    
%    \IEEEauthorblockA{\IEEEauthorrefmark{2}UC Berkeley, Department of Civil and Environmental Engineering} %
%    
%    \IEEEauthorblockA{\IEEEauthorrefmark{3}UC Berkeley, Department of Electrical Engineering and Computer Sciences} %
%    
%    \IEEEauthorblockA{\IEEEauthorrefmark{4}UC Berkeley, Department of Mechanical Engineering} %
%
%    \IEEEauthorblockA{\IEEEauthorrefmark{5}UC Berkeley, Institute for Transportation Studies}
}

% FOR IEEE T-RO SUBMISSION
% Our paper, titled "Framework for Control and Deep Reinforcement Learning in Traffic" will be presented at the IEEE Intelligent Transportation Systems Conference (ITSC) in 2017.

% \markboth{IEEE Transactions on Robotics,~Vol.~6, No.~1, January~2020}%
% \markboth{IN REVIEW (DO NOT DISTRIBUTE)}%
% \markboth{}%  % FOR ARXIV
\markboth{IEEE Transactions on Robotics}{}
% {Wu \MakeLowercase{\textit{et al.}}: Flow: A Modular Learning Framework for Mixed Autonomy Traffic}

\maketitle
\begin{abstract}
\newCathyTwo{
The rapid development of \textit{autonomous vehicles} (AVs) holds vast potential for transportation systems through improved safety, efficiency, and access to mobility. However, the progression of these impacts, as AVs are adopted, is not well understood.
% Due to numerous technical, political, and human factors challenges, new methodologies are needed to design vehicles and transportation systems for these positive outcomes. 
Numerous technical challenges arise from the goal of analyzing the partial adoption of autonomy: partial control and observation, multi-vehicle interactions, and the sheer variety of scenarios represented by real-world networks.
To shed light into near-term AV impacts, this article studies the suitability of deep \textit{reinforcement learning} (RL) for overcoming these challenges in a low AV-adoption regime.
A modular learning framework is presented, which leverages deep RL to address complex traffic dynamics. Modules are composed to capture common traffic phenomena (stop-and-go traffic jams, lane changing, intersections).
Learned control laws are found to improve upon human driving performance, in terms of system-level velocity, by up to 57\% with only 4-7\% adoption of AVs.
Furthermore, in single-lane traffic, a small neural network control law with only local observation is found to eliminate stop-and-go traffic -- surpassing all known model-based controllers to achieve near-optimal performance -- and generalize to out-of-distribution traffic densities.
}

\end{abstract}

\IEEEpeerreviewmaketitle

\enlargethispage{-3\baselineskip}  % Shrink bottom margin of current column

% \vspace{15mm}

\section{Introduction}
\label{sec:introduction}

Autonomous vehicles (AVs) are projected to enter society in the very near future, with full adoption in select areas expected as early as 2050~\cite{Wadud2016}.
However, the uncertainty in potential impacts is vast.
A recent study estimated that fuel consumption in the U.S. could decrease as much as 40\% or \textit{increase} as much as 100\% once autonomous fleets of vehicles are rolled out~\cite{Wadud2016}, potentially exacerbating the 28\% of energy consumption that is attributed to transportation in the US~\cite{US2016}.
% These factors include incorporation of platooning and eco-driving practices, vehicle right-sizing, induced demand, travel cost reduction, and new mobility user groups.
As such, computational tools are needed for the design, study, and control of these complex, large-scale robotic systems.

%\todoCathy{Consider adding the numbers for potential fuel and time saved as well.} 

Existing tools are largely limited to two commonly studied regimes: where AVs are few enough as to not affect the surrounding traffic dynamics~\cite{thrun2006stanley, buehler2009darpa, Spieser2014}, or so ubiquitous as to become a coordination problem~\cite{Dresner2008, Horn2013, miculescu2019polling}.
For clarity, we refer to these as the \textit{isolated autonomy} and \textit{full autonomy} cases, respectively.
At the same time, the intermediate regime, which is the long and arduous transition from no (or few) AVs to full adoption, is poorly understood. 
We term this intermediate regime \textit{mixed autonomy}.
The understanding of mixed autonomy is crucial for the design of suitable vehicle controllers, efficient transportation systems, sustainable urban planning, and public policy in the advent of AVs.
This article focuses on autonomous vehicles, which we expect to be among the first robotic systems to enter and widely affect existing societal systems.
Additional highly anticipated robotic systems, which may benefit from similar techniques as presented in this article, include aerial vehicles, household robotics, automated personal assistants, and additional infrastructure systems.

\enlargethispage{-3\baselineskip}  % Shrink bottom margin of current column

\begin{figure*}[t]
\begin{multicols}{3}
  \includegraphics[width=\linewidth]{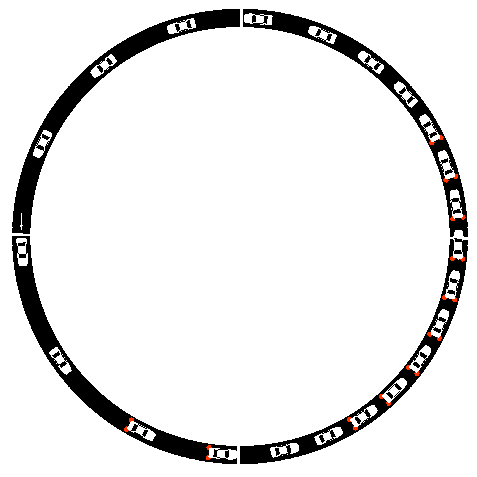} \par 
  \includegraphics[width=\linewidth]{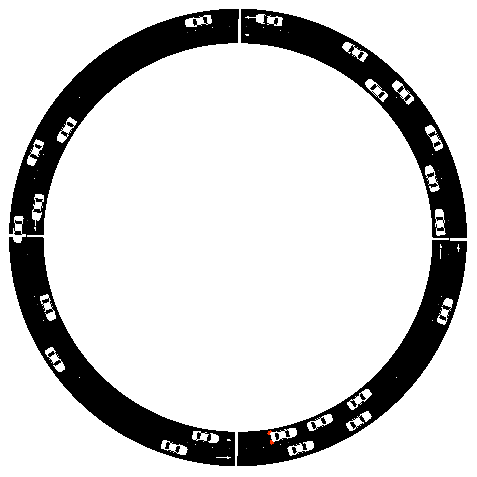} \par
  \includegraphics[width=\linewidth]{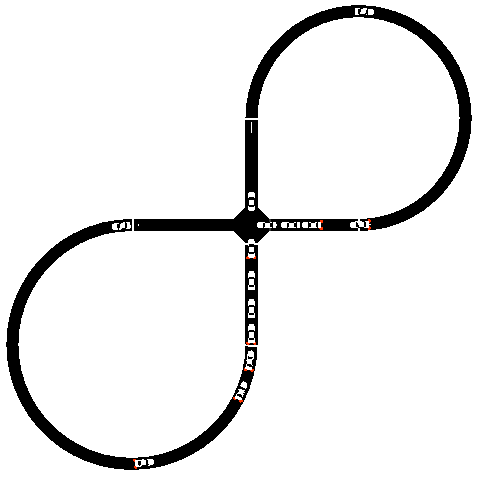}\par
\end{multicols}
\begin{multicols}{3}
  \includegraphics[width=\linewidth]
  {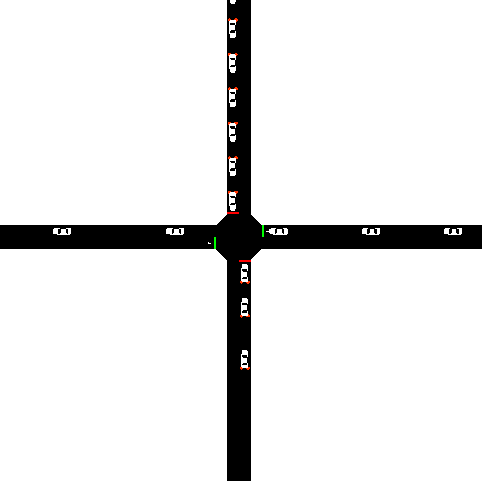} \par
  \includegraphics[width=\linewidth]{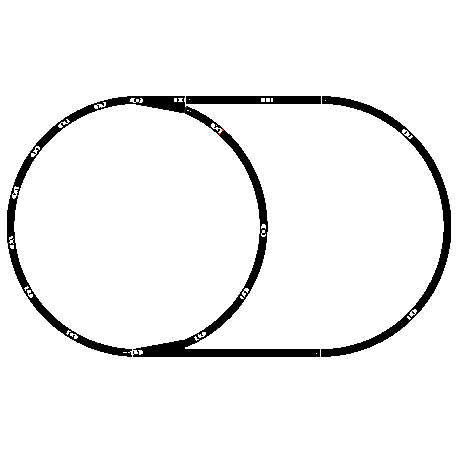} \par
  \includegraphics[width=\linewidth]{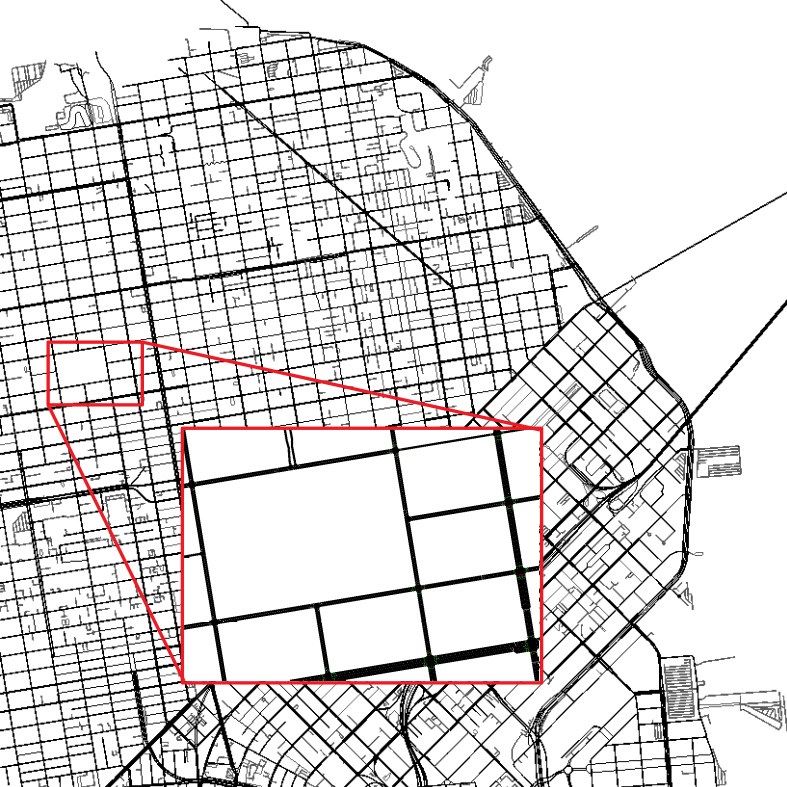} \par
\end{multicols}
\caption{\footnotesize \newCathy{Example network modules supported by the Flow framework. \textbf{Top left}: Single-lane circular track. \textbf{Top middle}: Multi-lane circular track. \textbf{Top right}: Figure-eight road network. \textbf{Bottom left}: Intersection network. \textbf{Bottom middle}: Closed loop merge network. \textbf{Bottom right}: Imported San Francisco network. In Flow, scenarios can be generated using OpenStreetMap (OSM) data and vehicle dynamics models from the traffic microsimulation package SUMO.}}
\label{fig:flow-networks}
\vspace{-10pt}
\end{figure*}

% Motivated by the importance and uncertainty about the \textit{transition} in the adoption of AVs, this article deviates from focus in the literature on isolated and full autonomy settings and proposes the mixed autonomy setting.
The mixed autonomy setting exposes heightened complexities due to the interactions of numerous human and robotic agents in highly varied contexts, for which the predominant analytical approaches of the traffic community are largely unsuitable.
Instead, we observe that model-free deep \textit{reinforcement learning} (RL) permits the decoupling of the mathematical modeling of the system dynamics from the control law design, thereby side-stepping limitations of classical approaches. % to autonomous vehicle control in complex environments.
Specifically, we propose a modular learning framework, in which environments representing complex control scenarios are comprised of reusable components, analogous to ``LEGO'' blocks.
We validate the proposed methodology on a classic traffic control scenario exhibiting backward-propagating traffic shockwaves in a partially-observed environment, and we subsequently produce a learned control law which far exceeds all previous methods, generalizes to unseen traffic densities, and closely matches theoretical performance bounds.
Finally, we demonstrate the efficacy of the framework by studying more complex traffic scenarios for which control-theoretic results are not known.
By appropriately composing the reusable components, we further demonstrate the effectiveness of the methodology with preliminary results on novel multi-lane, multi-AV, and intersection control scenarios.
% These results thereby show that deep reinforcement learning has the potential to produce new insights in mixed autonomy traffic, and in some cases, may even find near-optimal control laws.
To facilitate future research in mixed autonomy traffic, we developed and open-sourced the modular learning framework as Flow\footnote{Flow is open-source: \texttt{\url{https://github.com/flow-project/flow}}.}, which exposes design primitives for composing traffic control scenarios.
% \todoEugene{Think there's some kind of grammar bug in the sentence below, particularly around the line ",but can be arbitrary code". }
% \todoAboudy{replaced code with models, and added "analytically tractable" to differentiate the two things}
Our contributions aim to enable the community to study not only mixed autonomy scenarios which are composites of analytical mathematical frameworks, but also arbitrary networks or even full-blown traffic microsimulators, designed to simulate hundreds of thousands of agents in complex environments (see examples in Figure~\ref{fig:flow-networks}).
% Flow is a general framework which may also be of independent interest for the study of other forms of automation in traffic, including problems concerning traffic lights, road directionality, signage, roadway pricing, and infrastructure communication.

\todoCathy{Make clear in the intro that the reusable modules are components of a scenario and not the full scenarios themselves. Also be consistent with the terms reusable, modular, modules, etc.}

The contributions of Flow to the research community are multi-faceted. For the robotics community, Flow seeks to enable characterizations and empirical study of complex, large-scale, and realistic multi-robot control scenarios. For the machine learning community, Flow seeks to expose to modern RL algorithms to a class of challenging control scenarios derived from an important real-world domain. For the control community, Flow seeks to provide intuition, through successful learned-control laws, for new provable control techniques for traffic-related control scenarios. Finally, for the transportation community, Flow seeks to provide a new methodological pathway, through reusable traffic modules and modern RL methods, that addresses new challenges concerning AVs and long-standing challenges concerning traffic control.
% \todoEugene{Love the above paragraph, please keep it!}
% \todoAboudy{Ugh, fine.}

% END new intro

The rest of the article is organized as follows:
\newCathy{Section~\ref{sec:mixed-autonomy} introduces the problem of mixed autonomy.
Section~\ref{sec:related work} presents related work to place this article in the broader context of automated vehicles, traffic flow modeling, and deep RL.
Section~\ref{sec:background} summarizes requisite concepts from RL and traffic.
Section~\ref{sec:overview} describes the modular learning framework for the scalable design and experimentation of traffic control  scenarios.
% Section~\ref{sec:networks} presents the various building blocks used by SUMO for building general networks (underlying maps).
% Section~\ref{sec:taskSpace} presents the various settings for the optimization, incl. action / observation space, reward functions and control laws.
This is followed by two experimental sections: Section~\ref{sec:mixed-autonomy-ring}, in which we validate the modular learning framework on a canonical traffic control scenario; and Section~\ref{sec:flow-experiments} which presents more sophisticated applications of the framework to more complex traffic control scenarios.
}

\section{Mixed autonomy}
\label{sec:mixed-autonomy}

% To shed light into the integration of AVs into transportation systems, 
To shed light into the progression of impacts that AVs may have as they are adopted into transportation systems,
this article introduces the problem of \textit{mixed autonomy}: that is, given a traffic system with a fraction $p$ of AVs, what level of performance is achievable under utility function $U_p(\cdot)$?
% to frame the study of partial adoption of autonomy into an existing system.
% Although mixed autonomy technically subsumes isolated and full autonomy, both are still important and active areas of research and will often lead to more efficient algorithms for specialized settings.
% However, we observe that there are important robotics problems that are not addressed by either body of approaches, and thus formulate the problem of mixed autonomy traffic and propose a suitable methodology.
More generally, for a particular type of \textit{autonomy}, or advanced automation, (e.g. AVs, traffic signals, roadway pricing), mixed autonomy is the intermediate regime between a system with no adoption of autonomy and a system where autonomy is fully employed.
For example, in the context of AVs, full autonomy corresponds to 100\% of vehicles being autonomously driven, no autonomy corresponds to 0\% AVs, and mixed autonomy corresponds to some fraction $p$ of vehicles being AVs.
In this article, we take $U_p(\cdot)$ to be the average velocity of the system.

\renewcommand{\arraystretch}{1.5}
\begin{table}[th!]
\scriptsize
\centering
\begin{tabular}{ ll|l|l|l|l|l|l|l|l|l|l|l|l|l|l|l|l|l|l|l|l|l|l|l|l| } 
 \cline{3-4}
                      & & \multicolumn{2}{c|}{Uncertainty in system dynamics} \\
%                      & & \multicolumn{2}{c|}{\multirow{2}{*}{Uncertainty in system dynamics}} \\
%                      & & \multicolumn{2}{c|}{} \\
 \cline{3-4}
                     & & Low & High \\
 \hline
 \multicolumn{1}{|c|}{\multirow{2}{*}{Uncertainty in objective}} & Low & -- & Isolated autonomy \\
 \cline{2-4}
 \multicolumn{1}{|c|}{} & High & Full autonomy & Mixed autonomy \\
 \hline
\end{tabular}\\[10pt]
\caption{Uncertainties arising in autonomy settings.}
\label{table:mixed-autonomy-problem}
\vspace{-10pt}
\end{table}

Mixed autonomy differs from the earlier introduced cases in two nuanced but important ways, summarized in Table~\ref{table:mixed-autonomy-problem}.
First, the evaluation criteria of interest in mixed autonomy settings may exhibit more uncertainty than in isolated autonomy.
Isolated autonomy is often evaluated with respect to a known outcome, such as human or expert performance for a similar task.
The canonical example is the performance of a single AV, compared to a human driver~\cite{Bojarski2016, ShalevShwartz2016}.
Similarly, expert demonstrations are often considered a gold standard in robotic learning for a variety of tasks, including locomotion, grasping, and manipulation~\cite{atkeson1997robot, abbeel2004apprenticeship, Levine2016}.
In these cases, we assume both knowledge of a good control law and that it is feasible to attain. %, e.g. from a human or expert demonstrator.
However, evaluating with respect to a known outcome makes implicit assumptions about the capabilities of the autonomous system and the optimality of the known outcome, both of which may be incorrect.
For example, in the context of traffic networks, evaluating with respect to the known human performance is restrictive; it is well known that human driving behavior induces (suboptimal) stop-and-go traffic in a wide regime of traffic scenarios~\cite{Sugiyama2008, Orosz2010}.
\todoEugene{Do you need to define stop and go traffic before you use it as a standard term?}
In mixed autonomy settings, we are instead interested in evaluating with respect to a broader performance measures, such as traffic congestion, to understand potential effects on the system as a whole; the measures may be system-level, may be unattainable, and may be only partially controllable. % largely beyond the control of autonomy.
% The evaluation with respect to human performance allows for understanding whether algorithms can indeed achieve human-level performance on a task.

Second, the degree of uncertainty in the \textit{system dynamics} is greater in mixed autonomy than in full autonomy.
In contrast to isolated autonomy, full autonomy is indeed often evaluated with respect to a system-level objective, such as average travel time.
However, its system dynamics often exhibit low uncertainty
% much simpler than that of mixed autonomy
due to the simple fact that much of the system's state evolution is directly determined by autonomy.
That is, all autonomous components are known, as are the effects of their actions; and uncertainty from human behavior, common to isolated and mixed autonomy, is largely eliminated.
The low uncertainty in system dynamics permits direct modeling and analysis within a number of powerful mathematical frameworks, including partial differential equations~\cite{Lighthill1955, Richards1956, payne1979freflo, aw2000resurrection}, ordinary differential equations~\cite{gipps1981behavioural, Bando1994, Bando1995, Treiber2000}, and queuing systems~\cite{miller1961queueing, heidemann1996queueing, van2007modeling}.
In a few cases, control-theoretic performance bounds or optimal controllers can even be analytically derived. % or an optimal controller may even be found directly.
Even so, full autonomy is far from solved.
% ; however, numerous full autonomy settings can be analyzed using techniques from control theory, in particular when the system dynamics may be modeled 
In contrast, mixed autonomy suffers from additional challenges including interactions with humans of unknown or complex dynamics, partial observability from sensing limitations, and partial controllability due to the lack of full autonomy.
Therefore, a strict coupling between the mathematical modeling and the system evaluation is not sensible for studying mixed autonomy; in this article, we thus take a model-free approach.
% However, these control theoretic performance bounds be employed as reference points when evaluating mixed autonomy systems.
% For mixed autonomy settings, these aspects are often prohibitively complex to characterize within a single mathematical framework.

Mixed autonomy thus inherits the challenges from both isolated autonomy and full autonomy (see Table~\ref{table:mixed-autonomy-problem}).
In this work, we propose and validate a new methodology for addressing mixed autonomy in the context of traffic networks.
We posit that sampling-based optimization allows us to decouple mathematical modeling of the system dynamics and control-law design for arbitrary evaluation objectives, thereby overcoming the limitations of studies in both isolated and full autonomy.
In particular, we propose that model-free deep RL is a compelling and suitable framework for the study of mixed autonomy.
The decoupling allows the designer to specify arbitrary control objectives and system dynamics to explore the effects of autonomy on complex systems.
For the system dynamics, the designer may model a system of interest in whichever mathematical or computational framework they wish, and we require only that the model is consistent with a (Partially Observed) Markov Decision Process ((PO)MDP) interface.
For control law design, deep neural network architectures may be used for representing large and expressive control law (also called \textit{policy}) classes.
Finally, the resulting framework employs model-free deep RL to enable the designer to explore the effects of autonomy on a complex system, up to local optimality with respect to the control law parameterization.

\section{Related Work}
\label{sec:related work}

\noindent \textbf{Control of automated vehicles.} 
Automated and autonomous vehicles have been studied in a myriad of contexts; here, we describe prior work in isolated, full, and mixed autonomy.

\textit{Isolated autonomy}. Spurred by the US DARPA challenges in autonomous driving in 2005 and 2007~\cite{Thrun2005, buehler2009darpa}, countless efforts have demonstrated the increasing ability of vehicles to operate autonomously on real roads and traffic conditions, without custom traffic infrastructure.
These vehicles instead rely largely on on-board sensors (LIDAR, radar, camera, GPS), computer vision, motion planning, mapping, and behavior prediction, and are designed to obey traffic rules.
Robotics has continued to demonstrate tremendous potential in improving transportation systems through AV research; highly related problems include localization~\cite{Sukkarieh1999, Dissanayake2001, Cui2003}, path planning~\cite{Shiller1991, Bopardikar2015}, collision avoidance~\cite{Minguez2009}, and perception~\cite{Kanatani1990}.
Considerable progress has also been made in recent decades in vehicle automation, including anti-lock braking systems (ABS), adaptive cruise control (ACC), lane keeping, lane changing, parking, overtaking, etc.~\cite{Drakunov1995, VanArem2006, Lee2014, Son2015, Lefevre2014, Hatipoglu2003, Corporation1934, Paromtchik1996, Milanes2012}, which also have great potential to improve safety and efficiency in traffic.
The development of these technologies is currently focused on the performance of the individual vehicle, rather than its interactions or effects on other parts of the transportation system.

% The problems addressed in this article are part of a broader core set of robotics challenges concerning the deployment of multi-agent automation systems, such as fleets of self-driving cars as seen in~\cite{Pavone2012,Wu2017d}, coordinated traffic lights~\cite{Xie2012, Belletti2018}, or other coordinated infrastructure.

\textit{Full autonomy}.
At the other end of the autonomy spectrum, all vehicles are automated and operate efficiently with collaborative control.
Model-based approaches to such \textit{full autonomy} have permitted the reservation system design and derivation of vehicle weaving control for fully automated intersections~\cite{Dresner2008, miculescu2019polling} and straight roads~\cite{Horn2013}.

%\todoKathy{Too general; Alex's comment: maybe write in a way that relates to what we do: 'automated operations' as one of the building blocks of this future}
%\todoCathy{Add citations.}

%~\cite{Stevens} used reinforcement learning on traffic lights to increase traffic flow through intersections. 
% http://cs229.stanford.edu/proj2016spr/report/047.pdf
% \todoCathy{This work is not published as far as I can tell, so we shouldn't cite it.}

\textit{Mixed autonomy}.
A widely deployed form of vehicle automation is adaptive cruise control (ACC), which adjusts the longitudinal dynamics for vehicle pacing and driver comfort, and is grounded in classical control techniques~\cite{TechnicalCommitteeISO/TC2042010, Ioannou1993, Vahidi2003}.
Similarly, cooperative ACC (CACC) uses control theory to simultaneously optimize the performance of several adjacent vehicles, such as for minimizing the fuel consumption of a vehicle platoon~\cite{Shladover2005, Rajamani2002, Lu2004, Sheikholeslam1992, VanArem2006}.
Partial (C)ACC adoption is a form of mixed autonomy.
This article similarly studies longitudinal vehicle controllers, with a key difference being our focus on system-level rather than local objectives, such as the average velocity of all vehicles in the system, which is important for system operations and planning.

% However, there is no such result for a mixed autonomy intersection.

A few studies have started to use formal techniques to design controllers for system-level evaluation of mixed autonomy traffic, including state-space~\cite{Cui2017} and frequency domain analysis~\cite{wu2018stabilizing}.
There are also several modeling- and simulation-based evaluations of mixed autonomy systems~\cite{Kesting2007a, Yuan2009, Au2014} and model-based approaches to mixed autonomy intersection management~\cite{sharon2017protocol}.
Despite these advances in controller design, these approaches are generally limited to simplified models, such as homogeneous, non-delayed, deterministic driver models, or restricted network configurations.
% The present article presents the first model-agnostic study of system-level optimization of mixed autonomy traffic through modern machine learning techniques.
This article proposes to overcome these barriers through model-agnostic sampling-based methods. %, for the important reason that these disparate traffic dynamics exist simultaneously in real world systems and thus new approaches are needed to rigorously analyze the full systems.
Impressively, concurrent work by Stern, et al.~\cite{stern2018dissipation} demonstrated, in field operational tests, a reduction in fuel consumption of 40\% by the insertion of an autonomous vehicle in traffic in a circular track to dampen the famous instabilities displayed by Sugiyama, et al.~\cite{Sugiyama2008}.
% These tests motivate the present work: it demonstrates the power of automation and its potential impact on complex traffic phenomena such as \textit{stop-and-go} waves~\cite{PicolliBook}. 

On a large-scale network, fleets of AVs have been studied for shared-mobility systems, such as autonomous mobility-on-demand~\cite{Pavone2012, Spieser2014, Zhang2014}, which abstracts out the low-level vehicle dynamics and considers a queuing theoretic model.
Low-level vehicle dynamics, however, are crucial~\cite{Sadigh2016} because many low-level traffic phenomena affect energy consumption, safety, and travel time~\cite{Sugiyama2008, Lee2016, Rios2017a, Rios2017b}.
% In some settings, model-based controllers enable analytical solutions, or tractable algorithmic solutions. However, often, due to the nonlinearity of the models, numerous guarantees are lost in the process of developing controllers (i.e. optimality, run-time, complexity, approximation ratio, etc.).
% For example, while the ring setting enables elegant controllers to work in practice, the extension of these results (both theoretical and experimental) to arbitrary settings (network topologies, number of lanes, heterogeneity of the fleet, etc.) is challenging.
%These simulations produce interesting results at the system level. However, to move RL-based controllers from simulation to field experiments, simulation fidelity and existing vehicle and AI technologies must be addressed, among other issues. %Flow simulates perfect worlds most ideal for understanding system-level dynamics. However, reality is much more messy and requires more granular modeling and simulation of phenomena down to the level of vehicle dynamics. 

% \subsection{Background}

\addspace
\noindent \textbf{Single-lane traffic.} 
Although single-lane traffic has been studied for decades, the focus has been on modeling and control for local performance (e.g. comfort), rather than system-level performance (e.g. traffic congestion).
Therefore, we take this to be our starting point.
Various modeling approaches include closed networks~\cite{Sugiyama2008, Orosz2010, Cui2017, zheng2018smoothing}, open networks~\cite{Ioannou1993, Liang2000, wu2018stabilizing}, different human driving models~\cite{treiber2013traffic}, and different objectives~\cite{Kesting2007a}, 
% Relatively little work has been able to broach the problem of how to address undesirable traffic congestion through the control of vehicles.
To the best of the authors' knowledge,   work has achieved an optimal controller in the mixed autonomy setting for single-lane traffic congestion. % (in the form considered in this article or its many variants).
While studies in eco-driving practices provide heuristic guidance to drivers to ease traffic congestion~\cite{CIECA2007, Barth2009}, the characterization of optimality of these practices has received limited attention.

The most closely related work includes Horn, et al.~\cite{Horn2013} and Stern, et al.~\cite{stern2018dissipation}.
Horn, et al.~\cite{Horn2013} presents a near-optimal controller for the full autonomy setting.
% Respectively, these works do not achieve near-optimality and do not consider the mixed autonomy setting.
%TODO: reference EMERGENT BEHAVIORS paper as early work. Emergence of tailgating behavior.
Stern, et al.~\cite{stern2018dissipation} presents two hand-designed control laws for the mixed autonomy setting, which incorporate knowledge of the environment and thus we refer to them as model-based control laws. % of 21 human-driven vehicles and one vehicle employing one of two proposed AV control laws, which 
These control laws are included in our experiments as baseline methods and are detailed in Appendix~\ref{sec:controllers}.
% Importantly, both works hand-designed a control law based on knowledge of the environment, and thus we refer to them as model-based control laws.
In contrast, the approach proposed by this work requires significantly less ``design supervision'' in the form of a reward function, which avoids explicitly employing domain knowledge or mathematical analysis.
% which we detail in Sections~\ref{sec:stopper-follower}~and~\ref{sec:pi-controller}.
% Wu, et al.~\cite{Wu2017d} built upon an earlier version of this manuscript~\cite{wu2017flow}, and characterized the emergence of behaviors in mixed autonomy traffic. % , rather than on learning an optimal controller.

% \todoCathy{Consider adding a comparison with having 1 bilateral controller and 21 IDM vehicles. In that case, we would want to compare the time-to-stability. Maybe wait for the next pass on this.}

\todoEugene{There's a figure Alex often uses where the ring is depicted with the observed vehicles in blue, the AV in red, and the humans in white. I think this might be helpful to include?}

% \todoAboudy{While more complex mixed-autonomy settings can be studied in the context of deep RL and Flow, the present article emphasizes the performance of these frameworks on controlling a single lane ring road because it represents the forefront of traffic control with field-tested actuated devices. These field-tested results are the byproduct of decades of extensive numerical and analytical studies on the core factors leading to the formation of congestion in single lane ring road settings~\cite{herman1959traffic, orosz2006subcritical, Orosz2010}, with potentially similar advances and domain-specific experience required for future operational tests in more complex network geometries (e.g. open settings involving merges, lane drops, etc.). By demonstrating the comparative performance of control strategies derived from deep RL methods, the present article aims to accelerate the field of mixed-autonomy traffic control by promoting the use of these techniques in designing controllers for a plethora of more complex traffic control settings, as has been recently achieved in similar robotics tasks~\cite{levine2016end}.}

\addspace
\noindent \textbf{Modeling and control of traffic.}
Mathematical modeling and analysis of traffic dynamics is notoriously complex and yet is a prerequisite for traffic control~\cite{treiber2013traffic,Papageorgiou2003}, regardless of whether the control input is a vehicle, traffic light, ramp meter, or toll.
Such mathematical frameworks include partial differential equations, ordinary differential equations (ODEs), queuing systems, and stochastic jump systems; for the modeling of highway traffic, longitudinal dynamics, intersections, and lateral dynamics, respectively.
Researchers trade the complexity of the model (and thus its realism) for the tractability of analysis, with the ultimate goal of designing optimal and practical controllers. % such as safety or comfort~\cite{Ioannou1993,Vahidi2003,TechnicalCommitteeISO/TC2042010,Martin2013}. 
Consequently, results in traffic control can largely be classified as simulation-based numerical analysis with rich models but minimal theoretical guarantees~\cite{Liang2000, Bose2003, Ioannou2005, Kamal2014}, or theoretical analysis on simplified models such as assuming non-oscillatory responses~\cite{Swaroop1997} or focusing on a single-lane \newCathy{circular track}~\cite{Orosz2010, Orosz2011, Jin2014, Horn2013, Wang2016, wu2018stabilizing}.

% For instance, in ACC models, controllers and conditions for linear stability have been proposed for suppressing stop-and-go waves~\cite{Swaroop1994,Liang1999,Liang2000}.
% CACC additionally permits the model-based design of compressive platoons, similar to those studied in this article.
% Approaches to vehicle controller design include model predictive control for steering control~\cite{Falcone2007predictive, Falcone2008Mpc} and traffic control with automated vehicles~\cite{Baskar2009, Wang2014Rolling, Kamal2014}, and frequency domain analysis~\cite{Naus2010, Jin2014}.

Analysis techniques are often tightly coupled with the mathematical framework. For instance, linear systems theory techniques may be paired with ODEs and stochastic processes may be paired with queuing systems, but they may be incompatible with other mathematical frameworks.
Thus, there are virtually no theoretical works that simultaneously study lateral dynamics, intersection dynamics, longitudinal dynamics, etc., for the reason that they are typically all modeled using different mathematical frameworks.
As a notable exception, the work of Miculescu and Karaman~\cite{miculescu2019polling} takes a model-based approach which considers both intersections and longitudinal dynamics for a two-way fully-automated intersection with simplified dynamics.
This article seeks to take a step towards decoupling the reliance of control from the mathematical modeling of the problem, by proposing abstractions which permit the composition of reusable modules to specify problems and optimize for locally optimal controllers.

% For instance, for longitudinal control (forwards-backwards driving), such as for adaptive cruise control (ACC)~\cite{TechnicalCommitteeISO/TC2042010, Ioannou1993, Vahidi2003} and cooperative ACC (CACC)~\cite{Shladover2005, Rajamani2002, Lu2004, Sheikholeslam1992, VanArem2006}, typically optimize local metrics such as driver comfort or local fuel consumption.

% With the advent of autonomous vehicles, new frameworks and techniques are urgently needed to establish a foundation for studying the complex control and the effects of autonomous vehicles, thereby preparing the world for their adoption.
% Modern reinforcement learning techniques indicate promise towards the goal of obtaining controllers with desirable (though perhaps not optimal) properties while simultaneously studying complex settings.
\eat{\todoCathy{Thin down the above related works; move some of them into the section where we compare against controllers.}}

% \subsection{Vehicle controller design}

% FROM IEEE-TRO-v1 intro

%\todoCathy{From Alex (comment 9): "Assuming this is going to a robotics journal, somewhere we need a statement and some references (ask Steve Shladover or the new student I will introduce you next) about marching the valley of death between microsim and physical experiment. This needs to come early in the paper as a justification as to why we are doing this for a robotics journal." I'm going to punt this until we have Saleh's take on physical experiments.}

\addspace
\noindent \textbf{Deep reinforcement learning (RL).}
Deep RL is a powerful methodology for sequential decision making~\cite{bertsekas1996neuro, Sutton1998}, which inherits from machine learning and optimal control, and has demonstrated success in complex, data-rich problems such as Atari games~\cite{Mnih2015}, 3D locomotion and manipulation~\cite{Schulman2016a,Schulman2015a,Heess2015}, and navigation of stratospheric balloons~\cite{bellemare2020autonomous}.
% chess~\cite{Lai2015}, % chess~\cite{Lai2015},
% among others.
Deep RL is the key workhorse in our framework.
The advances in deep RL provide a promising alternative to model-based controller design.
More broadly, analyzing mixed autonomy traffic with RL is an intermediate step towards eventual deployment of autonomous fleets, providing insight to the system designer % of the vehicle controller or the transportation system
about a variety of complex performance metrics.
\addspace
\noindent \textbf{Deep RL and Traffic.}
% Several recent studies incorporated ideas from deep learning in traffic optimization.
Deep RL has been used for traffic prediction~\cite{Polson2016, Lv2015} and control~\cite{Li2016, Belletti2018, wei2019colight}.
A deep RL architecture was used by Polson, et al.~\cite{Polson2016} to predict traffic flows, demonstrating success even during special events with nonlinear features. To learn features to represent states involving both space and time, Lv, et al.~\cite{Lv2015} additionally used hierarchical autoencoding for traffic flow prediction.
Deep Q Networks (DQN) were employed for learning traffic signal timings in Li, et al.~\cite{Li2016}.
A multi-agent deep RL algorithm was introduced in Belletti, et al.~\cite{Belletti2018} to learn a control law for ramp metering.
Wei, et al.~\cite{wei2018intellilight, wei2019colight} employs RL and graph attention networks for control of traffic signals.
For additional uses of deep learning in traffic, we refer the reader to Karlaftis, et al.~\cite{Karlaftis2011}, which presents an overview comparing non-neural statistical methods and neural networks in transportation research.
These results demonstrate the promise of deep RL for traffic problems.
This article is the first to employ deep RL to design controllers for AVs and assess their impacts on traffic flow.
An early prototype of Flow is published~\cite{Wu2017a} and an earlier version of this manuscript is available~\cite{wu2017flow}.
In comparison, this article provides a substantive presentation of the mixed autonomy problem, the learning framework, case studies, and experimental findings which contribute to the understanding of AVs and traffic dynamics.

%~\cite{Stevens} used reinforcement learning on traffic lights to increase traffic flow through intersections. 
% http://cs229.stanford.edu/proj2016spr/report/047.pdf
% \todoCathy{This work is not published as far as I can tell, so we shouldn't cite it.}
% \todoCathy{I would like the related works to be about 1 page total. Any suggestions on what to cut?}
% \todoAboudy{I commented out a few redundant examples. We can consider doing more of that, if Eugene's suggestions don't cover everything.}

\section{Preliminaries}
\label{sec:background}
\label{sec:preliminaries}

\eat{Now, we describe two well-studied topics, which are key to Flow: longitudinal dynamics~\cite{Orosz2010} and reinforcement learning~\cite{Sutton1998}.
Longitudinal dynamics describe the forwards-backwards control of vehicle control models. Markov decision processes is the problem framework under which reinforcement learning methods optimize policies.}

We now define notation and key concepts used subsequently.

\subsection{Markov Decision Processes}

The framework described in this article tackles scenarios which conform to the standard interface of an episodic finite-horizon discounted \textit{Markov decision process} (MDP)~\cite{Bellman1957, Howard1964}, defined by the tuple $(\SS, \AA, P, r, \rho_0, \gamma, T)$, where $\SS$ is a (possibly infinite) set of states, $\AA$ is a (possibly infinite) set of actions, $P: \SS \times \AA \times \SS \to \R_{\geq0}$ is the transition probability distribution, $r: \SS \times \AA \to \R$ is the reward function, $\rho_0: \SS \to \R_{\geq 0}$ is the initial state distribution, $\gamma \in (0, 1]$ is the discount factor, and $T$ is the time horizon for an episode. Partially observable scenarios conform to the interface of a \textit{partially observable Markov decision process} (POMDP), and two more components are required: $\Omega$, a set of observations, and $\OO: \SS \times \Omega \to \R_{\geq 0}$, the observation probability distribution.

\textit{A note on terminology}: We use the term \textbf{scenario} to describe a full traffic setting, including all learning and non-learning components. It conforms to a (PO)MDP interface.
In robotics, the term ``task'' is commonly used; we prefer the word "scenario" to also capture settings with no learning components (e.g., situations with only human driver models).

Although traffic may be most naturally formulated as an infinite horizon problem, traffic \textit{phenomena} such as traffic jams are ephemeral or even periodic. Thus, we formulate finite horizon MDPs.
More generally, traffic has periodic patterns on a daily or weekly basis. Thus, the finite horizon problem can be a suitable approximation of the infinite horizon problem, so long as the horizon is sufficiently long to capture the transient or periodic behavior. For example, in the single-lane track scenario (Section~\ref{sec:mixed-autonomy-ring}), the periods are around 40 sec. To enable the formation of the periodic behavior, we select a fairly long horizon length of 300 seconds (or 3000 simulation steps).
% In our case of the circular track, we selected a 300 second simulation duration (6 minutes), because even the worst of the stop-and-go waves could be dissipated in that amount of time (see Figure~4).
% The scenarios presented in this article, the double-lane track and Figure 8 networks exhibit periodic behaviors. 
Furthermore, we will select a high discount factor (close to 1) to approximate a non-discounted problem.
% Another related critique may concern the choice of studying a discounted MDP, rather than an average cost MDP. In principle we are as concerned with the speed of traffic now as we are in future timesteps. Similarly, although many real-world problems in robotics in actuality should be studied in an infinite horizon average cost context, deep RL methods are largely designed for discounted MDPs, which have an effective horizon of $\mathcal{O}(1/(1-\gamma))$. In principle, one can approximate a true infinite horizon problem by taking the discount factor $\gamma \to 1$; for this reason, we use a large discount factor $\gamma = 0.999$.

\subsection{Reinforcement learning}

RL studies the problem of how agents can learn to take actions in its environment, often formulated as an (PO)MDP, to maximize its cumulative reward~\cite{bertsekas1996neuro, Sutton1998}.
This article uses policy gradient methods~\cite{sutton2000policy}, a class of RL algorithms which optimize a stochastic policy $\pi_\theta : \SS \times \AA \to \R_{\geq 0}$, e.g. deep neural networks.
Although commonly called a policy, we will generally refer to $\pi$ as a controller or control law in this article, to be consistent with traffic control terminology.
% For a stochastic policy, both the mean and standard deviation are predicted by the controller. Actions are then sampled from a corresponding Gaussian distribution. During test time, the mean action is taken, corresponding to the maximum likelihood deterministic controller.
% Policy gradient methods iteratively update the parameters of the control law by estimating a gradient for the expected cumulative reward, using data samples (e.g. from a traffic simulator or model).
Three classes of control laws are considered: \textit{Linear network}, \textit{Multilayer Perceptron} (MLP), and \textit{Gated Recurrent Unit} (GRU).
The Linear network is a parameterized linear function.
The MLP is a classical artificial neural network with one or more hidden layers~\cite{haykin1994neural}, consisting of linear weights and nonlinear activation functions (e.g. \texttt{tanh}, ReLU).
The GRU is a recurrent neural network that makes use of parameterized update and reset gates, which enable decision making based on both current and past inputs~\cite{chung2015gated}.
% capable of storing information from previous states~\cite{chung2015gated}.
% GRUs make use of parameterized update and reset gates, which enable decision making based on both current and past inputs. % and are also optimized by the policy gradient method.
% In all cases, the networks are trained using backpropagation to optimize its parameters.

% It is designed to incorporate memory in the network; i.e., it maintains a hidden internal state. This enables the network to make decisions based on both current input and past inputs.

\subsection{Vehicle dynamics models}
% \subsection{Traffic system dynamics}

The environments studied in this article are traffic systems.
Basic traffic dynamics on single-lane roads can be represented by \textit{ordinary differential equation} (ODE) models known as \textit{car following models} (CFMs).
% \noindent\textbf{Car following models:}
% Note that, for consistency with notation commonly used in the literature of car following models, this subsection uses notation self-contained and used only in Section~\ref{sec:car-following-models}.
These models describe the longitudinal dynamics of human-driven vehicles, given only observations about itself and the vehicle preceding it.
CFMs vary in terms of model complexity, interpretability, and their ability to reproduce prevalent traffic phenomena, including stop-and-go traffic waves.
% We now describe standard CFMs and the concept of equilibrium velocity.
% In the next section, we detail a commonly used CFM called the intelligent driver model, which we employ for the longitudinal dynamics of the human drivers in the numerical experiments of this article.
For modeling of more complex traffic dynamics, including lane changing, merging, driving near traffic lights, and city driving, we refer the reader to the text of Treiber and Kesting~\cite{treiber2013traffic} dedicated to this topic.

Standard CFMs are of the form:
\begin{equation}
    a_i = \dot v_i = f(h_i, \dot h_i, v_i),
    \label{eq:cfm}
\end{equation}
where the acceleration $a_i$ of car $i$ is some typically nonlinear function of $h_i, \dot h_i, v_i$, which are the headway, relative velocity, and velocity for vehicle $i$, respectively.  Though a general model may include time delays, % from the input signals $h_i, \dot h_i, v_i$ to the resulting output acceleration $a_i$, 
we will consider a non-delayed system, where all signals are measured at the same time instant.  Example CFMs include the Intelligent Driver Model (IDM)~\cite{Treiber2000} and the Optimal Velocity Model (OVM)~\cite{Bando1994, Bando1995}. IDM is used in the experiments of this article to model human driving (see Appendix~\ref{sec:idm}).

% START reusable modules

\section{Flow: A modular learning framework}
% \section{De-coupling the Scenario and the Controller} % Flow-verview :)
\label{sec:overview}
\label{sec:modules}

\begin{figure*}[!th]
\centering
\includegraphics[width=0.95\textwidth]{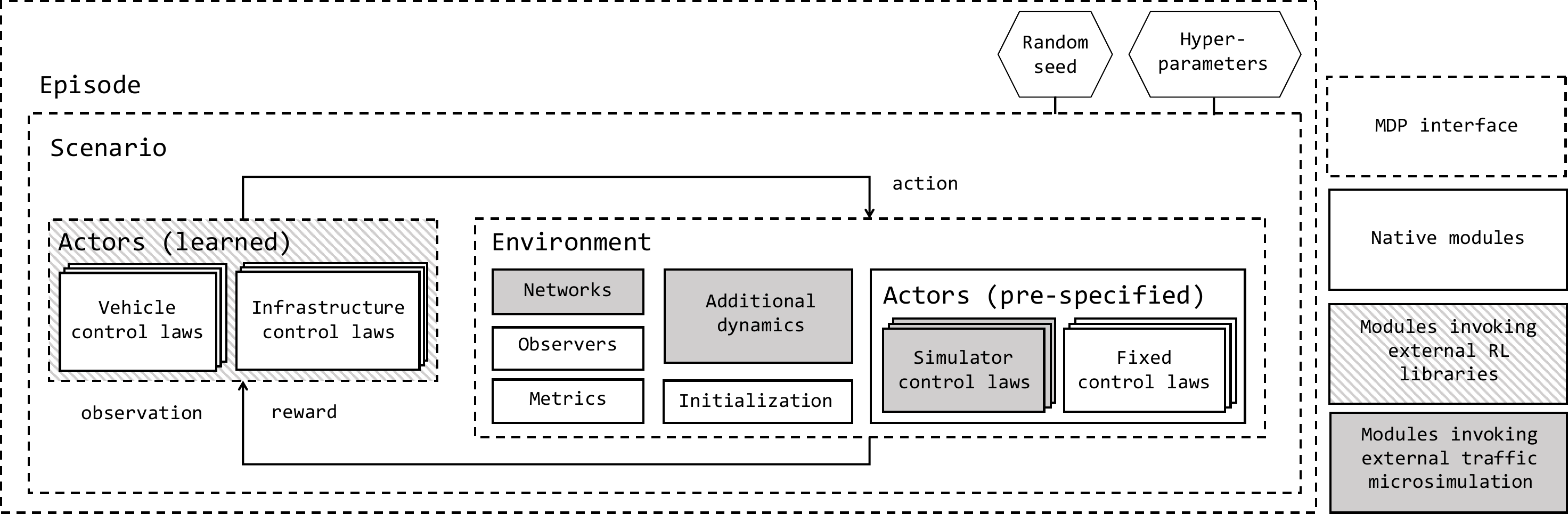}
\caption{\footnotesize \newCathy{Flow is a modular learning framework, which enables the composition of diverse mixed autonomy traffic scenarios for study with deep reinforcement learning.
Scenarios conform to a (PO)MDP interface (dotted rectangles).
Modules (solid rectangles) of varying types are composed to form scenarios. % Module types include networks, actors, observers, control laws, metrics, initializations, and additional dynamics.
Actors may include vehicles or infrastructure, and may be learned or pre-specified.
Additional dynamics may include vehicle fail-safes, right-of-way rules, and physical limitations.
Additional parameters (hexagons) may also be configured.
% Flow facilitates the composition of these modules to create new traffic scenarios of interest, which may then be studied as traffic control problems using deep RL methods.
Flow invokes external libraries for RL training and simulation.}}
\label{fig:flow-overview} % Flow-verview :)
\vspace{-10pt}
\end{figure*}

% \subsection{\newCathy{Reusable modules for configurable traffic scenarios}}

\todoEugene{To beat a dead horse dead, I think this part of the paper emphasizing how this is the first really modular RL env and how RL is actually really incredible for complicated, modular envs is really good! We should emphasize these ideas more in the intro.}

% While the research community in deep RL focuses on solving pre-defined tasks, relatively little emphasis is placed on \newCathyTwo{flexible} scenario creation (see Section~\ref{sec:related work}).
While RL testbeds have enabled significant progress for algorithm development~\cite{Bellemare2013, Todorov2012, Brockman2016}, testbeds which enable RL to shed light into important real-world problem domains remain limited.
% To this end, this article contributes a modular learning framework 
% The ability to create modular environments for traffic control is limited.
Moreover, in multi-agent systems such as traffic, the flexibility to study a wide range of scenarios is important due to their varied and complex nature, including different numbers and types of vehicles, heterogeneity of agents, network configurations, regional behaviors and regulations, etc.
To this end, this article contributes an approach that decomposes a scenario into modules which can be configured and composed to create new scenarios of interest.
Flow is the resulting modular learning framework for enabling the creation, study, and control of complex traffic scenarios.

% This complexity motivates our design of reusable modules, which can be easily pieced together to produce traffic scenarios of interest.
% These reusable modules... \todoCathy{one-sentence summary of what they are.}
% Flow is a modular learning framework, which exposes these reusable modules to ease designing of configurable traffic scenarios and learning of the subsequent control laws.
% Specifically, Flow employs deep reinforcement learning as a learning framework for the controller design.
% however, other frameworks such as genetic algorithms, POMDP solvers, Monte Carlo Tree Search, and local search may be employed as well.
% Flow additionally utilizes a traffic microsimulator in the construction of several modules.

% \todoCathy{@aboudy I am renaming scenario $\to$ network in the article, as discussed. We can discuss if it's time to make the change in the codebase.}
% \todoAboudy{I added the issue as a task for flow-0.5.0}

\subsection{Scenario modules}
\label{sec:modules}

Flow is comprised of the following modules, which can be assembled to form traffic scenarios of interest (see Figure~\ref{fig:flow-overview}).

% \todoEugene{I think it might be helpful to ground some of these by providing a quick example for some of them. For example, observer and control laws might be improved by adding that an example of an observer would be returning the states of the vehicles nearby the AV etc. }

\smallskip \noindent \textbf{Network:}
The network specifies the physical road layout, in terms of roads, lanes, length, shape, roadway connections, and additional attributes.
Examples include a two-lane circular track with circumference 200m, or the structure described by importing a map from OpenStreetMap (see Figure~\ref{fig:flow-networks} for examples of supported networks).
Several of these examples will be used in demonstrative experiments in Sections~\ref{sec:mixed-autonomy-ring}~and~\ref{sec:flow-experiments}.
More details about the specific networks are in Appendix~\ref{sec:networks}.
% Based on the specifications provided, the net and configuration files needed by SUMO are generated. The user also specifies the number and types of vehicles (car following model and a lane-change controller), which will be placed in the scenario.

\smallskip \noindent \textbf{Actors:} The actors describe the physical entities in the environment which issue control signals to the environment.
In contrast to isolated autonomy settings, in a traffic setting, there are typically many interacting physical entities.
Due to the focus on mixed autonomy, this article specifically studies vehicles as its physical entities.
Other possible actors may include pedestrians, bicyclists, traffic lights, roadway signs, toll booths, as well as other transit modes and infrastructure.

\smallskip \noindent \textbf{Observer:}
The observer describes the mapping $\mathcal{S} \to \mathcal{O}$ and yields the function of the state that is observed by the actor(s).
The output of the observer is taken as input to the control law, described below.
For example, while the state may include the position, lane, velocity, and accelerations of all vehicles in the system, the observer may restrict access to only information about local vehicles and aggregate statistics, such as average speed or queue length at an intersection.

\smallskip \noindent \textbf{Control laws:}
Control laws dictate the behaviors of the actors and are functions mapping observations to control inputs $\mathcal{O} \to \mathcal{A}$.
All actors require a control law, which may be pre-specified or learned.
For instance, a control law may represent a human driver, an autonomous vehicle, or even a set of vehicles.
That is, a single control law may be used to control multiple vehicles in a \textit{centralized control} setting.
Alternatively, a single control law may be used by multiple actors in a \textit{shared parameter control} setting.
% For instance, all human drivers may follow the same control law.

\smallskip \noindent \textbf{Dynamics:}
The dynamics module consists of additional submodules which describe different aspects of the system evolution, including vehicle routes, demands, stochasticity, traffic rules (e.g., right-of-way), and safety constraints.

\smallskip \noindent \textbf{Metrics:}
The metrics describe pertinent aggregated statistics of the environment.
The reward signal for the learning agent is a function of these metrics.
Examples include the average velocity of all vehicles and the number of hard braking events.

\smallskip \noindent \textbf{Initialization:}
The initialization describes the initial configuration of the environment at the start of an episode.
Examples include setting the position and velocity of vehicles according to different probability distributions.

\smallskip
Sections~\ref{sec:mixed-autonomy-ring}~and~\ref{sec:flow-experiments} demonstrate the potential of the framework.
Whereas in a model-based framing, many modules are simply not re-configurable due to differences in the mathematical descriptions (e.g. discrete versus continuous control inputs, such as in the case of longitudinal and lateral control), in this model-agnostic framework, disparate dynamics may be captured in the same scenario and effectively studied using sampling-based optimization techniques such as deep RL.
 
% END reusable modules

%\subsection{Architecture}
%\label{sec:architecture}

%\begin{figure}[!th]
%\centering
%\includegraphics[width=0.47\textwidth]{figures/CISTAR-updated-diagram.png}
%\caption{\footnotesize Flow Architecture. A Flow experiment involves a scenario and environment, interfaced with rllab or Ray RLlib and controllers. The experiment scenario runs a generator to create road networks for use in SUMO, which is started by the environment. Controllers and rllab or Ray RLlib take experiment states and return actions, which are applied through SUMO's TraCI API. (See Section~\ref{sec:flow-architecture}).}
%\label{fig:flow-architecture}
%\end{figure}

% An experiment using Flow requires defining two components: a scenario and an environment. These and several supporting components as well as their interactions are summarized in \figref{flow-architecture}.

% Based on the specifications provided, the net and configuration files needed by SUMO are generated. The user also specifies the number and types of vehicles (car following model and a lane-change controller), which will be placed in the scenario.

\subsection{Architecture and implementation}
\label{sec:architecture}

% Although the architecture is agnostic to specific machine learning and traffic software packages, we chose to integrate widely used open-source tools to promote access and extension.

The implementation of Flow is open source and builds upon open source software to promote access and extension.
The project aims to support the development of custom modules and thereby permit the study of richer and more complex environments, agents, metrics, and algorithms.
The implementation builds upon SUMO (\textbf{S}imulation of \textbf{U}rban \textbf{MO}bility)~\cite{Krajzewicz2012} for traffic modeling, Ray RLlib~\cite{liang2017ray} for RL methods, and OpenAI gym~\cite{Brockman2016} for the MDP interface.
% \todoEugene{This is a bit confusing to me, SUMO is not a continuous time traffic simulator? It uses an Euler/ballistic update step which is an approximation to continuous time systems?}
SUMO is a microscopic traffic simulator, which explicitly models individual vehicles, pedestrians, traffic lights, and public transportation.
It supports urban-scale road networks. % and of modeling the dynamics of each vehicle in the simulation.
Flow utilizes SUMO's Python API, TraCI (\textbf{Tra}ffic \textbf{C}ontrol \textbf{I}nterface).
Ray RLlib is a distributed framework for training and evaluating of RL algorithms. %  on a variety of scenarios, from classic tasks such as cartpole balancing to more complicated tasks such as 3D humanoid locomotion.
OpenAI gym is an MDP interface for RL tasks.
% The library provides an easy-to-use suite of reinforcement learning tasks.

% \todoEugene{Would it be worthwhile to describe/provide an example of what defining an MDP entails e.g. set up an example of states, actions and rewards}

Flow is implemented as a lightweight architecture to connect the modules described in the previous section and permit experimentation.
As typical in RL, an \textbf{environment} encodes the MDP (scenario).
The environment facilitates the composition of dynamics and other modules, stepping through the simulation, retrieving the observations, sampling and applying actions, computing the metrics and reward, and resetting the simulation at the end of an episode. 
% The environment is updated at each timestep of the simulation and, importantly, stores each vehicle's state (e.g. position and velocity). Information from the environment is provided to a controller or passed to rllab or Ray RLlib to determine an action for a vehicle to apply, e.g. an acceleration. Note that the amount of information provided to either RL or to a controller can be restricted as desired, thus allowing fully observable or partially observable MDPs. This article studies both fully and partially observed settings.
\newCathy{A \textbf{generator} produces network configuration files compatible with SUMO according to the network description.
The generator is invoked by the \textbf{experiment} upon initialization and, optionally, upon reset of the environment, allowing for a variety of initialization conditions, such as sampling from a distribution of vehicle densities.
% The experiments presented in this article include large loop roads generated by specifying the number of lanes and ring circumference, figure-eight networks with a crossing intersection, closed loops with merging networks, and standard intersections.
Flow then assigns control inputs from the different control laws to the corresponding actors, according to an \textbf{action assigner}, and uses the TraCI library to apply actions for each actor.}
Actions specified as accelerations are converted into velocities, using numerical integration and based on the timestep and the current state.

\eat{ Because TraCI can only set velocities, not accelerations, we convert the acceleration into an instantaneous $\delta v =  \alpha \cdot dt$, where $\alpha$ is the acceleration and $dt$ is the simulation time step-size. }
\eat{\todoKanaad{Let's bring up a point about mulitagent Flow work here somewhere? I'm not completely sure what that would entail? do we want to say like "Flow facilitates shared policies by training the same policy using multiple vehicles to sample trajectories."?}}

% Flow is a computational framework for traffic microsimulation with RL methods.
% Flow thereby seeks to fill the gap between modern machine learning and complex control problems in traffic.

Finally, Flow is designed to be inter-operable with classical model-based methods for evaluation purposes.
In other words, the learning component of Flow is optional, and this permits the fair comparison of diverse methods for traffic control.
% Together, this permits the study of mixed method, diverse, heterogeneous, and mixed autonomy settings.
% Examples and further implementation details are provided in Appendix~\ref{sec:controllers}~and~\ref{sec:fail-safes}, respectively, including information about supported controllers and fail-safe mechanisms.

\eat{
\section{Experiment setup}
\label{sec:experiment-setup}
\todoKathy{"Do you want to rename 'Experiment Framework'?}

\todoCathy{Note: This section is for setup configurations that are common among many of the experiments in this paper.}

We evaluate the performance of state of the art deep reinforcement learning techniques on a variety of traffic scenarios, which are described below, and, when possible, compare them to established control-theoretic results.
}

\section{Configurable modules for mixed autonomy}
\label{sec:mixed-autonomy-ring}
\label{case-study:mixed-autonomy-ring}

% Through the composition of the modules presented in Section~\ref{sec:modules} and the application of modern deep reinforcement learning methods, 
This section demonstrates that deep RL can solve a classic yet challenging traffic scenario.
Specifically, the canonical setup of Sugiyama, et al.~\cite{Sugiyama2008} is studied, which consists of 22 human-driven vehicles on a circular track with a circumference of 230 m.
This seminal experiment shows that human driving causes backward propagating traffic waves, resulting in some vehicles to come to a complete stop (see left side of Figures~\ref{fig:stabilizing-the-ring-vel}~and~\ref{fig:mixed-ring}).
Remarkably, this occurs even in the absence of typical sources of traffic perturbations, such as lane changes, merges or stop lights.
To analyze the impact of AVs, we adapt the setup to mixed autonomy using the framework presented in Section~\ref{sec:modules}.

\subsection{Experiment Modules}
We design the following experiment, in which one human driver is replaced by an AV, by composing modules proposed in Section~\ref{sec:modules}. The network and simulation-specific parameters of the numerical experiments are summarized in Table~\ref{table:network-simulation}.

% \todoEugene{Unformly sampled from 220 to 270? I think it'd be helpful to point out very clearly somewhere why you do this. It's scattered through the article that RL works across across a wide range of densities, but there isn't one place where it's very clearly stated what the experiment is and why it's set up the way it is.}
\smallskip \noindent \textbf{Network:} \newCathyTwo{The training domain consists of single-lane circular track networks (Figure~\ref{fig:flow-networks}, top left), with uniformly sampled track lengths $L \sim \text{Unif}([L_{\text{min}}, L_{\text{max}}])$}. Selecting a range of tract lengths represents a continuous range of traffic conditions and avoids overfitting to a single track length. We take $L_{\text{min}} = 220$ m and $L_{\text{max}} = 270$ m.
An alternative approach (not considered here) to represent a range of traffic densities is to fix the track length and instead vary the number of vehicles.
% The variation on length represents a wide range of traffic densities.

\begin{table}[]
\centering
\begin{tabular}{ |l|l| } 
 \hline
 \textbf{Experiment parameters}     & \textbf{Value}   \\
 \hline
 simulation step         & 0.1 s/step       \\ 
 circular track range (train) & [220, 270] m            \\ 
 circular track range (test) & [210, 290] m            \\ 
 warmup time             & 75 s             \\
 time horizon            & 300 s            \\
 total number of vehicles      & 22               \\
 number of AVs      & 1               \\
 \hline
\end{tabular}
\caption{\newCathy{Network and simulation parameters for mixed autonomy circular track (single-lane) experiment.}}
\label{table:network-simulation}
\end{table}

\smallskip \noindent \textbf{Actors:} There are $n=22$ vehicles, each 5 m long.

\smallskip \noindent \textbf{Observer:}
The state consists of the velocity and position of all vehicles, that is $s = (x_1, v_1, x_2, v_2,\dots, x_{n}, v_{n})$. It is practical to consider an observer which restricts the observation to the information that can be directly sensed by the single learning agent, given as $o = (\frac{v_i}{v_0}, \frac{\dot{h}_i}{v_0}, \frac{h_{i}}{L_{\text{max}}})$, where $i$ is the index of the learning agent, $v_0$ is the speed limit, $L_{\text{max}}$ is the maximum length of the track, headway $h_i = x_{i-1} - x_{i}$, and headway differential $\dot{h}_i = v_{i-1} - v_{i}$. The observation can be viewed as normalized inputs to a car-following model.
% The observer maps the full state to only the velocity of the autonomous vehicle, the relative velocity to its preceding vehicle, and its relative position to the preceding vehicle.
% \eat{ We additionally compare against another partially observed setting, where the density is additionally observed, and a fully observed setting, where all vehicle positions and velocities are observed.}

\smallskip \noindent \textbf{Control laws:}
% Separate control laws are used for the behavior of human drivers and AVs.
Of the 22 actors, 21 are modeled as human drivers \newCathyTwo{according to} the \textit{Intelligent Driver Model} (IDM)~\cite{treiber2013traffic} (see Appendix~\ref{sec:idm} and Table~\ref{table:controller-params}).
% When all 22 are modeled using human driver models, traffic jams are exhibited (see left side of Figures~\ref{fig:stabilizing-the-ring-vel}~and~\ref{fig:mixed-ring}), similarly to Sugiyama, et al.~\cite{Sugiyama2008}, and thereby serves as a \textit{no autonomy} baseline for comparison.
% Due to the sampling-based nature of the proposed methodology, the performance of previous methods may be readily assessed relative to the RL approach.
% This provides a type of backwards compatibility with the research community, and permits the fair comparison of control laws, regardless of the underlying mathematical framework.
For the single autonomous vehicle actor, we compare the following control laws.
Recall that actors \newCathyTwo{partially observe the environment.}

\noindent \textit{Learned control laws:}
\begin{itemize}
\item \newCathy{GRU (memory): hidden layer (5), \texttt{tanh} non-linearity.}
\item \newCathy{MLP (no memory): diagonal Gaussian MLP, two-layer network with hidden layers (3,3), \texttt{tanh} non-linearity.}
\item Linear network (no memory).
\eat{\item Learned agent with MLP control law, with full observability.}
\end{itemize}

\noindent \textit{Model-based control laws:}
\begin{itemize}
\item \textit{FollowerStopper}~\cite{stern2018dissipation} (see Appendix~\ref{sec:stopper-follower}), with desired velocity parameter fixed at 4.15 m/s, calibrated for a track length of 260 m;
% The \textit{FollowerStopper} control law is introduced in Stern, et al. and is also detailed in Section~\ref{sec:stopper-follower}.
% \textit{FollowerStopper} requires an external desired velocity, so we selected the largest fixed velocity which successfully mitigates stop-and-go waves at a track length of 260 m;
this is further discussed in the results. \eat{This variant sets the target velocity of the \textit{FollowerStopper} control law at the equilibrium velocity of the system. \todoCathy{Correct this to what we eventually do.}}
\item \textit{Proportional Integral} (PI) control with saturation, given in Stern, et al.~\cite{stern2018dissipation} and is detailed in Appendix~\ref{sec:pi-controller}.
\item IDM: For a \textit{no autonomy} baseline. % , we take IDM as the control law. This results in a \textit{stop-and-go} stable limit cycle. Although not strictly a lower bound, we take this to be a practical lower bound, since any worse controller could be replaced by human driving behavior.
\end{itemize}
% \item Human driver using the \textit{Intelligent Driver Model} (IDM), with partial observation, which is presented in more details in Appendix~\ref{sec:controllers}. This setting yields traffic jams as in~\cite{Sugiyama2008} and serves as a baseline comparison.

\smallskip \noindent \textbf{Dynamics:}
\newCathyTwo{The overall system dynamics consists of a cascade of nonlinear dynamics models from $n-1$ (homogeneous) actors and $1$ autonomous vehicle actor (learning agent).}
The \newCathyTwo{$n-1$ IDM dynamics models} are additionally perturbed by Gaussian acceleration noise of $\mathcal{N}(0,0.2)$, calibrated to match measures of stochasticity to the IDM model presented by Treiber, et al.~\cite{treiber2017intelligent}.
\newCathyTwo{The traffic simulator enforces safety through built-in failsafe mechanisms.}
Before starting the 300 second episode, there is a warmup period of 75 seconds, in which the acceleration of the AV is overridden by the IDM model to allow for randomization of the initial state and for the formation of stop-and-go waves.

% Using Flow, we study the control of the mixed autonomy single-lane ring road, which extends the canonical and widely studied problem of stabilizing the (homogeneous) single-lane ring road.
% We will show that the shortcomings of hand-designed controllers are exhibited by state-of-the-art controllers in the example of the mixed autonomy ring setting. 
% \todoCathy{@Alex how do we be more tactful here (in the sentence before)?}
% In this and following sections, we study the potential of RL to produce well-performing control laws, in highly nonlinear and complex settings.\eat{ We evaluate the performance of state of the art deep reinforcement learning techniques on a variety of traffic tasks, and compare them to established control results.}

\smallskip \noindent \textbf{Metrics:} We consider two natural metrics: the average velocity of all vehicles in the network and a control cost, which penalizes acceleration. The reward function supplied to the learning agent is a weighted combination of the two metrics.
\begin{equation}
\newCathyTwo{r(s,a) = \frac{1}{n} \sum_{i} v_i - \alpha |a|}
\end{equation}
\newCathyTwo{where $\alpha = 0.1$.}

\smallskip \noindent \textbf{Initialization:}
The vehicles are evenly spaced around the circular track, with an initial velocity of 0 m/s.

\subsection{Learning setup}
% \smallskip \noindent \textbf{\newCathy{Learning} setup:} 
The AVs execute parameterized control laws, trained using policy gradient methods.
We use the \textit{Trust Region Policy Optimization} (TRPO)~\cite{Schulman2015a} method, with linear feature baselines as described in Duan, et al.~\cite{Duan2016}, discount factor $\gamma = 0.999$, and step size 0.01.
% We consider control laws with and without memory.
% This specific example (single-lane traffic congestion benchmark) uses only a hidden layer of shape (3,3).
% For control laws with memory, we use a GRU with hidden layers (5,) and \texttt{tanh} non-linearity.
The numerical experiments were conducted on three Intel(R) Core(TM) i7-6600U CPU @ 2.60GHz processors for six hours. A total of 6,000,000 samples (167 driving hours) were simulated during the training procedure\footnote{Further implementation details can be found at: \texttt{\url{https://github.com/flow-project/flow-lab/tree/master/flow-framework}}.}.

\subsection{Performance bounds}
Before presenting the results, we discuss performance bounds, which provide a reference for evaluating the learned controllers.
Specifically, because we are concerned with \textit{steady-state performance} of the traffic system, we take the \textit{limit cycles} of the system to be the performance bounds.
Limit cycles are closed curves that trajectories tend towards, if stable (or away, if unstable).
Limit cycles generalize the notion of a system equilibria from a point to a trajectory.
% The overall traffic system is comprised of interconnected dynamical models, including CFMs, representing the many interacting vehicles in the traffic system.
% An example of a traffic system which can be represented by only longitudinal vehicle dynamics is a circle of $n$ vehicles, all driving clockwise, and each of which is following the vehicle preceding it.
% The result is a dynamical system which consists of a cascade of $n$ potentially nonlinear and delayed vehicle dynamics models.
% This example will serve as our first numerical experiment (Section~\ref{sec:mixed-autonomy-ring}), as it is able to reproduce important traffic phenomena (traffic waves and traffic jams), and, despite its simplicity, its optimal control problem in the mixed autonomy setting has remained an open research question.
% Although the primary methodology explored in this article is model-agnostic, we employ control theoretic model-based analysis to compute the performance bounds of the traffic system, so that we can adequately assess the performance of the learned control laws.
% In a homogeneous setting, where all vehicles follow the same dynamics model, 

\textit{Uniform flow} describes the situations where all vehicles move at some constant velocity $v^*$ and constant headway $h^*$, and corresponds to one of the limit cycles of the traffic system.
% Strictly speaking, this is an equilibria of the no autonomy setting, rather than the mixed autonomy setting.
For a general car following model $f$, the relationship between the equilibrium headway $h^*$ and equilibrium velocity $v^*$ is written as:
% They can be represented in terms of the vehicle dynamics model by:
\begin{equation}
  a_i = 0 = f(h^*, 0, v^*).
  \label{eq:uniformflow}
\end{equation}
% This equation defines For a general car following model $f$ the relationship between the equilibrium headway $h^*$ and equilibrium velocity $v^*$.
The specific relationship for IDM is displayed in Figure~\ref{fig:stabilizing-speed-density} (dotted green curve).
These equilibria have high velocities, which is desirable, but they are \textit{unstable} due to properties of human driving behavior~\cite{treiber2006delays}, and thus do not naturally occur.
We take this to be the performance upper bound.

% We use the term \emph{traffic condition} to refer to this equilibrium velocity $v^*$.
% It is intuitive to think of the equilibrium density (which is inversely related to the equilibrium headway $h^*$) as a traffic condition.
% We will seek to evaluate our learned control laws against a range of traffic conditions.
% Each traffic condition (density) has associated with it an optimal equilibrium velocity $v^*$.
% In practice, the equilibrium density can be approximated by the local traffic density.
% In settings with heterogeneous vehicle types, the equilibrium can be numerically solved by constraining the total headways to be the total road length and the velocities to be uniform.
% It is important to note the difference between the equilibrium velocity $v^*$ and the target velocity $v_0$ (free flow speed) of the vehicle models; $v_0$ can be thought of as a speed limit for highway traffic~\cite{Treiber2000}.
% On the other hand, $v^*$ is a control theoretic quantity jointly determined by the traffic condition $h^*$, the target velocity $v_0$, the system dynamics, and various other parameters.

On the other hand, a stable limit cycle of the system corresponds to \textit{traffic waves} (also called stop-and-go waves)~\cite{Orosz2010}.
% Another set of equilibria corresponds to \textit{traffic waves} (also called stop-and-go waves).
% These traffic waves are stable limit cycles, that is, closed curves that trajectories tend towards (rather than away)~\cite{Orosz2010}.
In other words, the traffic system tends towards traffic jams~(see Figure~\ref{fig:stabilizing-speed-density}, dotted red curve, for the resulting average velocity under IDM)).
This is a practical performance lower bound because any AV control law yielding worse performance could be replaced by a human driver model for a better outcome.
% Using IDM, the relationship between density and the average velocity with traffic waves is displayed in Figure~\ref{fig:stabilizing-speed-density} (dotted red curve).

% Finally, we are concerned with \textit{steady-state performance} of the traffic system, rather than any instantaneous performance.
% We thus take the uniform flow and traffic waves equilibria to describe the upper and lower bounds, respectively, for steady-state performance of the overall traffic system.
We note that, because we are analyzing the \textit{no autonomy} scenario, but evaluating in a \textit{mixed autonomy} scenario, these performance bounds should be viewed as close approximations to the true bounds.
For more detailed performance bounds, we refer the reader to related work~\cite{Orosz2010, Wu2017d, zheng2018smoothing}.

\subsection{Results}

% This benchmark study demonstrates that deep RL can be used to learn a controller which performs better than state-of-the-art model-based controllers in traffic scenarios.
% Through the study of a mixed autonomy single-lane ring, we demonstrate that Flow enables the fine-grained benchmarking of classical and learned controllers.
By studying the mixed autonomy track, we demonstrate 1) that Flow enables composing modules to study an open problem in traffic control and 2) that reliable controllers for complex problems can be efficiently learned, which surpass the performance of all known model-based controllers. % , such as controlling mixed-autonomy traffic (involving both autonomous and human-driven vehicles) in a ring road.
This section details our findings. Videos and additional results are also available\footnote{See videos: \texttt{\url{https://sites.google.com/view/ieee-tro-flow}}}.
\eat{\todoCathy{FINAL PASS: anywhere we say "compare" or "comparison", change it to "benchmark" or "benchmarking" (if it makes sense).}}

\setcounter{subsubsection}{0}
\subsubsection{Performance}
% \todoKathy{fix this section}
First, Figure~\ref{fig:stabilizing-speed-density} evaluates the AV controllers across a wide range of traffic conditions (210 to 290 m circumference tracks). % This traffic density versus velocity plot shows the performance of the different learned and model-based controllers.
We observe that GRU and MLP control laws match the optimal velocity closely for traffic densities, thereby practically eliminating congestion.
The PI with Saturation and \textit{FollowerStopper} control, on the other hand, only \newCathy{dissipate stop-and-go traffic} at densities less than or equal to their calibration density (less congested settings).
The Linear control law performs well but not as well as the MLP/GRU.
This indicates that a linear function may be unable to express the equilibrium flow velocity, whereas a two-layer neural network can.
Our learned controllers outperform all the model-based controllers, with the exception of the PI with saturation controller outperforming the Linear controller in low density traffic.

\begin{figure}[th]
\centering
\pgfplotstableread[col sep = comma]{data/avg_velocity_controllers.csv}\mydata
\begin{tikzpicture}
	\begin{axis}[
		legend style={nodes={scale=0.5, transform shape}},
		height=6cm, 
		width=9cm, 
		grid=major,
		ylabel=Average velocity (m/s), 
		ylabel near ticks,
		xlabel= Vehicle density (veh/km), 
		ymin=1.5,
		ymax=6.5,
		xmax=0104.761904761905, 
		xmin=0075.8620689655172,
        xtick distance=0005.,
		xticklabel style={
  			/pgf/number format/precision=3,
  			/pgf/number format/fixed
  		},
  		label style={font=\small},
  		tick label style={font=\small}
  	]

	\addplot [draw=none, mark=none, fill=gray, fill opacity=0.5, forget plot] coordinates { (0, 0) (0081.48, 0) (0081.48, 7) (0, 7)};
	\addplot [draw=none, mark=none, fill=gray, fill opacity=0.5, forget plot] coordinates { (0100, 0) (0200, 0) (0200, 7) (0100, 7)};
 
    \addplot[line width=0.5mm, black!30!green, dashed] table[x index = {0}, y index = {3}]{\mydata};
    \addlegendentry{Uniform flow unstable equilibrium (optimal)}

    \addplot[line width=0.4mm, blue] table[x index = {0}, y index = {6}]{\mydata};
    \addlegendentry{GRU control law (ours)}

    \addplot[line width=0.4mm, red!80!black] table[x index = {0}, y index = {7}]{\mydata};
    \addlegendentry{MLP control law (ours)}

    \addplot[line width=0.5mm, black!30!green] table[x index = {0}, y index = {8}]{\mydata};
    \addlegendentry{Linear control law (ours)}

    \addplot[line width=0.5mm, orange] table[x index = {0}, y index = {4}]{\mydata};
    \addlegendentry{PI with saturation control law}

    \addplot[line width=0.5mm, teal] table[x index = {0}, y index = {5}]{\mydata};
    \addlegendentry{FollowerStopper control law}

	\addplot[black, dotted] coordinates {(0084.6,0) (0084.6,7)};
    \addlegendentry{Calibration density for PI control law}
    
    \addplot[line width=0.5mm, red, dashed] table[x index = {0}, y index = {2}]{\mydata};
    \addlegendentry{Stop-and-go stable limit cycle}[fontsize=1cm]

	\end{axis}
\end{tikzpicture}
\caption{\footnotesize Performance of AV control laws for the single-lane mixed autonomy track.
The overall system velocity of learned (GRU, MLP, and Linear) and model-based (FollowerStopper and PI Saturation) control laws are averaged for the final 100 s of simulation time over ten runs at each evaluated density.
Also displayed are the performance upper and lower bounds, derived from the unstable and stable system limit cycles, respectively.
The white and gray regions indicate the training-time and testing-time densities, respectively.
% The GRU and MLP control laws closely match the velocity upper bound closely for all densities (train and test). \newCathyTwo{The Linear control law performs well but not as well as the MLP/GRU. The PI with Saturation and FollowerStopper baselines perform relatively well at their calibration density, but not as well as the learned controllers. The PI with saturation controller generalizes well to lower density traffic.} Remarkably, the GRU and MLP control laws are able to generalize \newCathyTwo{(to the gray region)} and bring the system to near optimal velocities even at densities outside the training range \newCathyTwo{(white region)}. Additionally, the learned MLP control law demonstrates that memory is not necessary to achieve near optimal average velocity.
}
% \todoCathy{Add variance / error bars for the 10 runs.}
% \todoAboudy{This'll probably make the figure look very cluttered, at best we can rewrite it as a table}
\label{fig:stabilizing-speed-density}
\end{figure}

\begin{figure*}[htbp]
\centering
\vspace{-20pt}
\setlength\fboxsep{0pt}
\setlength\fboxrule{0.25pt}
\pgfplotstableread[col sep = comma]{data/follower_stopper.csv}\mydata
\begin{tikzpicture}
	\begin{axis}[
		legend pos=south east,
		height=4cm, 
		width=16cm, 
		grid=major,
% 		title=Velocity Profile for Vehicles in a 260m Ring Road with a Follower Stopper Controller,
		title=Velocity Profile for Vehicles in a 260 m Circular Track for Different AV Control Laws,
		ylabel=velocity (m/s), 
		ylabel near ticks,
		xticklabel style={
  			/pgf/number format/precision=3,
  			/pgf/number format/fixed
  		},
  		ymin=-1,
  		ymax=11,
  		xmin=0,
  		xmax=600,
  		label style={font=\small},
  		tick label style={font=\small},
  		% to remove xticks
        x tick label style={color=white}
    ]

    \addplot[black!30!blue] table[x index = {0}, y index = {1}]{\mydata};
    \addlegendentry{FollowerStopper}[fontsize=1cm]

	\addplot[black, dashed, line width=1pt] coordinates {(300,-1) (300,11)};
	\addplot[black, <->, line width=1pt] coordinates {(5,10.5) (295,10.5)};
    \node[] at (axis cs: 150,9.7) {\small no automation (stop-and-go waves)};
    \addplot[black, <->, line width=1pt] coordinates {(305,10.5) (595,10.5)};
    \node[] at (axis cs: 450,9.7) {\small automation turned on (for velocity optimization)};
    \addplot[name path=A, draw=none] table[x index = {0}, y index = {2}]{\mydata};
    \addplot[name path=B, draw=none] table[x index = {0}, y index = {3}]{\mydata};
   	\addplot[draw=none, mark=none, fill=gray, fill opacity=0.4] fill between[
  		of = A and B,
  	];
  	
	\end{axis}
\end{tikzpicture}
\pgfplotstableread[col sep = comma]{data/pi_saturation.csv}\mydata
\begin{tikzpicture}
	\begin{axis}[
		legend pos=south east,
		height=4cm, 
		width=16cm, 
		grid=major,
% 		title=Velocity Profile for Vehicles in a 260m Ring Road with a PI Saturation Controller,
		ylabel=velocity (m/s), 
		ylabel near ticks,
		xticklabel style={
  			/pgf/number format/precision=3,
  			/pgf/number format/fixed
  		},
  		ymin=-1,
  		ymax=11,
  		xmin=0,
  		xmax=600,
  		label style={font=\small},
  		tick label style={font=\small},
  		% to remove xticks
        x tick label style={color=white}
]

	\addplot[black!30!blue] table[x index = {0}, y index = {1}]{\mydata};
    \addlegendentry{PI Saturation}[fontsize=1cm]
    
	\addplot[black, dashed, line width=1pt] coordinates {(300,-1) (300,11)};
    \addplot[name path=A, draw=none] table[x index = {0}, y index = {2}]{\mydata};
    \addplot[name path=B, draw=none] table[x index = {0}, y index = {3}]{\mydata};
  	\addplot[draw=none, mark=none, fill=gray, fill opacity=0.4] fill between[
  		of = A and B,
  	];
 
	\end{axis}
\end{tikzpicture}
\pgfplotstableread[col sep = comma]{data/mlp.csv}\mydata
\begin{tikzpicture}
	\begin{axis}[
		legend pos=south east,
		height=4cm, 
		width=16cm, 
		grid=major,
% 		title=Velocity Profile for Vehicles in a 260m Ring Road with a MLP Controller,
		ylabel=velocity (m/s), 
		ylabel near ticks,
		xticklabel style={
  			/pgf/number format/precision=3,
  			/pgf/number format/fixed
  		},
  		ymin=-1,
  		ymax=11,
  		xmin=0,
  		xmax=600,
  		label style={font=\small},
  		tick label style={font=\small},
  		% to remove xticks
        x tick label style={color=white}
    ]

    \addplot[black!30!blue] table[x index = {0}, y index = {1}]{\mydata};
    \addlegendentry{MLP control law (ours)}[fontsize=1cm]

	\addplot[black, dashed, line width=1pt] coordinates {(300,-1) (300,11)};
    \addplot[name path=A, draw=none] table[x index = {0}, y index = {2}]{\mydata};
    \addplot[name path=B, draw=none] table[x index = {0}, y index = {3}]{\mydata};
  	\addplot[draw=none, mark=none, fill=gray, fill opacity=0.4] fill between[
  		of = A and B,
  	];
 
	\end{axis}
\end{tikzpicture}
\pgfplotstableread[col sep = comma]{data/gru.csv}\mydata
\begin{tikzpicture}
	\begin{axis}[
		legend pos=south east,
		height=4cm, 
		width=16cm, 
		grid=major,
% 		title=Velocity Profile for Vehicles in a 260m Ring Road with a GRU Controller,
		ylabel=velocity (m/s), 
		ylabel near ticks,
        xlabel= time (s), 
		xticklabel style={
  			/pgf/number format/precision=3,
  			/pgf/number format/fixed
  		},
  		ymin=-1,
  		ymax=11,
  		xmin=0,
  		xmax=600,
  		label style={font=\small},
  		tick label style={font=\small}
    ]

    \addplot[black!30!blue] table[x index = {0}, y index = {1}]{\mydata};
    \addlegendentry{GRU control law (ours)}[fontsize=1cm]

	\addplot[black, dashed, line width=1pt] coordinates {(300,-1) (300,11)};
    \addplot[name path=A, draw=none] table[x index = {0}, y index = {2}]{\mydata};
    \addplot[name path=B, draw=none] table[x index = {0}, y index = {3}]{\mydata};
  	\addplot[draw=none, mark=none, fill=gray, fill opacity=0.4] fill between[
  		of = A and B,
  	];

	\end{axis}
\end{tikzpicture}
\caption{\textbf{Velocity profile for single-lane mixed autonomy track.} Sample evaluations start with 300 seconds where the AV is overridden by IDM human driving behavior to allow the formation of traffic waves, followed by 300 seconds with four different AV control laws. Both learned control laws bring the system to close to the 4.82 m/s uniform flow velocity. A successful evaluation of the PI Saturation controller is shown; however, it can be inconsistent in its performance across episodes. The \textit{FollowerStopper} falls short, settling at 4.15 m/s. The GRU control law reaches the optimal velocity fastest.}
% In addition, the \textit{FollowerStopper} controller is the most brittle and can only stabilize a 260 m ring road to a speed of 4.15 m/s
% \todoEugene{Is the y axis velocity or average velocitY?}
% \todoAboudy{Blue - velocity of AV, grey - upper and lower bounds of velocities of all vehicles}
\vspace{-10pt}
\label{fig:stabilizing-the-ring-vel}
\end{figure*}

Figure~\ref{fig:stabilizing-the-ring-vel} shows velocity profiles for the learned and model-based AV control laws on a 260 m track. Although both types of controllers eventually bring the system to uniform flow, the GRU control law reaches the equilibrium velocity fastest. The GRU and MLP control laws mitigate congestion with less oscillatory behavior than the \textit{FollowerStopper} and PI with Saturation control laws.
The \textit{FollowerStopper} control law is the least performant; it settles at a steady-state speed of 4.15 m/s, well below the 4.82 m/s equilibrium velocity.
% \todoCathy{Put these number into a table instead?}

\begin{figure*}[t]
\begin{multicols}{2}
% \centering
\includegraphics[width=0.49\textwidth]
{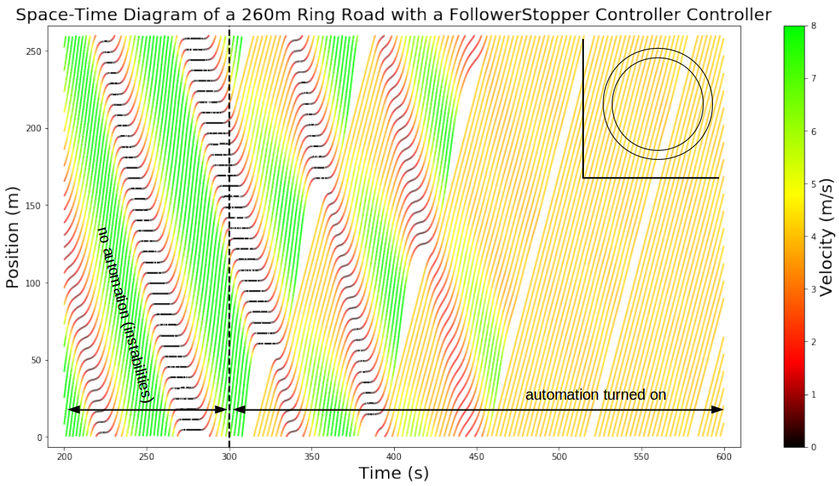}

\includegraphics[width=0.49\textwidth]{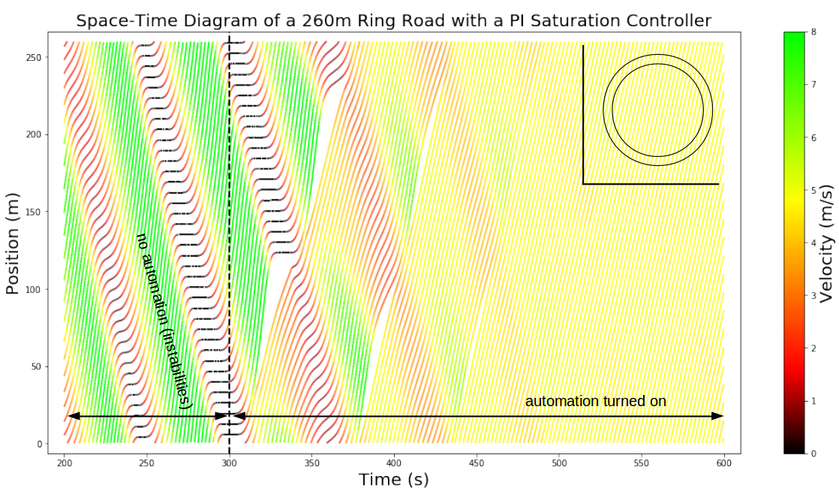}
\end{multicols}
\vspace{-8mm}
\begin{multicols}{2}
% \centering
\includegraphics[width=0.49\textwidth]{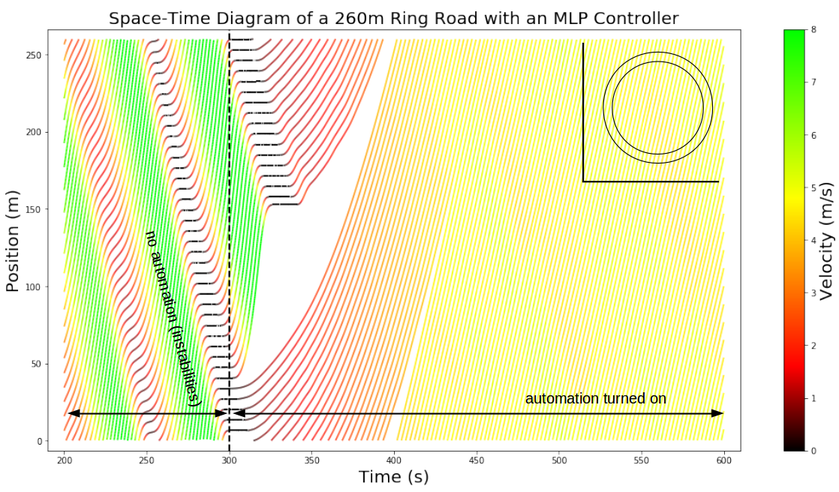}

\includegraphics[width=0.49\textwidth]{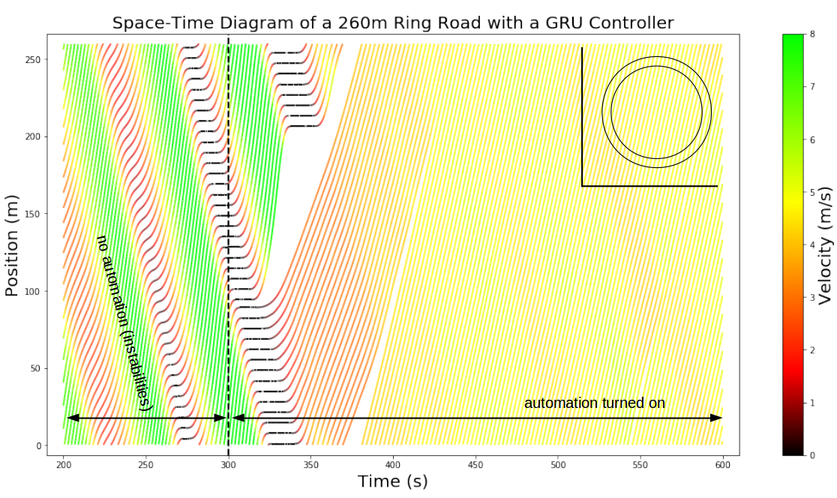}
\end{multicols}
\vspace{-5mm}
\caption{\footnotesize Space-time trajectories for single-lane mixed autonomy track, for learned and model-based AV control laws. Traffic waves form during the first 300 s.
% The AV then employs a model-based or learned control law.
%\textbf{Top left}: \textit{FollowerStopper} (model-based). \textbf{Top right}: PI with Saturation (model-based). \textbf{Bottom left}: Time-space diagram for an AV employing a learned MLP control law. \textbf{Bottom right}: Time-space diagram for an AV employing a learned GRU control law.
}
\label{fig:mixed-ring}
\vspace{-10pt}
\end{figure*}

Figure \ref{fig:mixed-ring} shows space-time curves for all vehicles, for different AV control laws. We observe that the PI with Saturation and \textit{FollowerStopper} control laws leave much smaller gaps (headways) than the MLP and GRU control laws. The MLP control law exhibits the largest gaps, as can be seen by the large white portion of the MLP plot within Figure~\ref{fig:mixed-ring}. If this had been a multi-lane scenario, then the smaller gaps would have the benefit of preventing opportunistic lane changes, so this observation can lead to improved reward design for more complex mixed autonomy traffic studies.

\eat{\todoCathy{Space-time curves for all 5 controllers (260m) -- column-size}
\todoCathy{Space-time curves for all PI saturation and GRU control law (230m) -- column-size, showing that their controller fails sometimes, whereas ours works.}
\todoCathy{For each controller: velocity vs time (260m) -- full width}
\todoCathy{Training: Comparison of the methods across different densities: velocity vs density -- column}}
\eat{
\todoCathy{Fundamental diagram??????}
\todoAboudy{I think we should forget the fundamental diagram idea}
}
%%%%%%%%%%%%%%%%%%%%%%%%%

% \todoKathy{Do we need to update the figure label?}

%%%%%%%%%%%%%%%%%%%%%%%%%%%%%%%%%%%%%%%%%%%%%%%%%%
% \begin{figure}[th]
% \centering
% \includegraphics[width=0.4\textwidth]{figures/follower_stopper_space_time.png}
%   \caption{Space time diagram for the ring road with a FollowerStopper Controller.}
% \label{fig:mixed-ring-follower-stopper}
% \end{figure}

% \begin{figure}[th]
% \centering
% \includegraphics[width=0.4\textwidth]{figures/pi_saturation_space_time.png}
%   \caption{Space time diagram for the ring road with a PI with Saturation Controller.}
% \label{fig:mixed-ring-pi-saturation}
% \end{figure}

% \begin{figure}[th]
% \centering
% \includegraphics[width=0.4\textwidth]{figures/mlp_space_time.png}
%   \caption{Space time diagram for the ring road with an MLP Controller.}
% \label{fig:mixed-ring-mlp}
% \end{figure}

% \begin{figure}[th]
% \centering
% \includegraphics[width=0.4\textwidth]{figures/gru_space_time.png}
%   \caption{Space time diagram for the ring road with a GRU Controller.}
% \label{fig:mixed-ring-gru}
% \end{figure}

\subsubsection{Robustness}
\label{sec:mixed-ring-robustness}

A strength of learned control laws is that they do not rely on external calibration of parameters that are specific to a traffic setting, such as traffic density.
On the other hand, in our experience, model-based controller baselines often exhibit considerable sensitivity to the traffic setting.
We found the performance of the PI with Saturation control law to be sensitive to parameters and initial conditions, even though in principle it adjusts to different densities with a moving average filter.
Using parameters calibrated for the 260 m track (as described in Stern, et al.~\cite{stern2018dissipation}), the control law performs decently at 260 m;
however, its performance quickly degrades at higher densities (more congested settings), dropping close to the performance lower bound \newCathyTwo{(Figure~\ref{fig:stabilizing-speed-density})}.
Additionally, even for a fixed track of length 260 m, it is inconsistent in mitigating the traffic waves.
A successful episode is shown in Figures~\ref{fig:stabilizing-the-ring-vel}~and~\ref{fig:mixed-ring}; however, unsuccessful episodes bring down its average performance (Figure~\ref{fig:stabilizing-speed-density}).

% \todoEugene{The strongest criticism I would make here, if I were a reviewer, is that you should have compared against a FollowerStopper where you adaptively updated the parameters as a function of a filtered estimate of the density. Might be an unfair thing to ask, but I could imagine Reviewer 2  suggesting this experiment?}
Similarly, the \textit{FollowerStopper} control law requires careful tuning before usage, which is beyond the scope of this work.
% suffers from the same calibration limitations as the PI with Saturation Controller.
Specifically, the desired velocity must be provided beforehand.
Interestingly, we found experimentally that this control law is often ineffective if provided too high of a desired velocity, even if it is well below the uniform flow equilibrium velocity.
% However, if a lower desired velocity is first provided as an intermediate control target, then the desired velocity may subsequently be achieved.

% This suggests that a simple control law such as the \textit{FollowerStopper} cannot optimally stabilize a mixed autonomy ring, and additionally, that there is additional tuning necessary to use the \textit{FollowerStopper} controller.

\subsubsection{Generalization of the learned control law}

% We observed that training with a range of vehicle densities encourages the learning of a more robust control law.
By training with a range of vehicle densities, we found the learned control laws to generalize even to densities outside of the training regime, leading to a more robust control law. Figure~\ref{fig:stabilizing-speed-density} shows the learned control laws closely tracking the performance upper bound in the testing regime.
Additionally, and interestingly, even training in the \textit{absence} of noise in the human driver models, learned control laws still successfully stabilized settings \textit{with} human model noise during test time (not shown).
\eat{\todoCathy{Add test-time velocity vs density plots, which show generalize-ability, testing each controller on a range outside of training (with both higher and lower densities) -- column-size.}}

\newCathy{
\subsubsection{Partial observability eases controller learning}

At this early stage of autonomous vehicle development, we do not yet have a clear picture of what manufacturers will choose in terms of sensing infrastructure for the vehicles, what regulators will require, or what technology will enable (e.g. communication technologies).
Furthermore, we do not know how the observation landscape of autonomous vehicles will change over time, as AVs are gradually adopted.
Therefore, a framework which is modular and provides flexibility for the study of AVs is crucial.
By invoking the composable observation components, we can readily study a variety of possible scenarios.

% In particular, policy gradient methods, restricted to use the same partial observation information of previously studied model-based controllers and with access to samples from the overall traffic system (via a black box simulator), achieves a near-optimal controller.
\eat{\todoCathy{Can we say something here about how the MLP policy without density information performs? Does it work at all (even for 1 density, but with randomization and noise)? Does it generalize at all to multiple densities?}
\todoCathy{What does it take to learn a controller as good as / better than the hand-designed controllers?}
\todoAboudy{@Alex We probably shouldn't blatantly say that we can outperform existing state-of-the-art controllers.}}

As such, this study considers a partially-observed setting for several reasons: 1) it is the more realistic setting for near-term deployments of autonomous vehicles, and 2) it permits a fair comparison with previously studied model-based controllers, which typically utilize partial observation.
% Furthermore, there are hand-designed controllers in the literature for this setting, with which we can benchmark.
Finally, since we found the learned control laws to achieve the optimal velocity curve, we do not extensively explore the use of fuller observations.
However, in Section~\ref{sec:flow-experiments}, we do explore a variety of additional settings, ranging from partially observed to fully observed settings.
% We would expect that the fully observed setting (with a MLP control law) would perform as well if not better than our learned control laws in the partially observed setting.

Our partially observed experiments uncover several surprising findings which warrant further investigation.
First, contrary to the classical view on partially observed control (e.g. POMDPs), these experiments suggest that partial observability may ease training instead of making it more difficult; as compared to full observations, we found that partial observations decreased the training time required from around 24 hours to 6 hours.
Second, as seen in Figure~\ref{fig:stabilizing-speed-density}, the results demonstrate that a near global optimum is achievable even under partial observation.
Finally, the MLP control law closely mirrors the GRU control law and the optimal velocity curve; despite the partially observed setting, this suggests that memory is not necessary to achieve near optimal velocity across the full range of vehicle densities with a single learned controller.
% \todoCathy{@aboudy. Would you happen to have numbers on this one? In my recollection, the partially observed ring takes like 1-2 hours and the fully observed one takes 9-10 hours? Let's try to provide some numbers on this counter-intuitive result.}
% \todoAboudy{Check some of the red comments. I wrote the partially observable case took six hours on my laptop. The fully observable is not a fair comparison, since it was a while ago and using different training batch sizes, etc.. But it was around 12-24 hours.}

A possible explanation is that a neural network with fewer weights may require fewer samples and iterations to converge to a local optimum, thus contributing to faster training.
A more rigorous understanding of this phenomenon is left as a topic of future study, as well as questions concerning the situations under which partial observations still lead to a globally optimal solution in a learning framework.
These early results suggest that deep RL methods may more efficiently utilize partial observations when they are provided appropriately, avoiding the need to learn to discount extraneous inputs.
}

% \subsection{Discussion}

% \noindent \textbf{Discussion:}

\eat{
TO ALEX: how much would it strengthen the paper to explore the fully observed setting?
\todoCathy{Consider including this result experimentally.}
\todoCathy{How well does MLP policy with full observability perform vs GRU policy? (Do we NEED recurrence?)}
\todoCathy{? Reference results of linearized controller analysis (our CDC submission) / Cui’s state-space analysis of Dan’s controllers?}
\todoCathy{Add a table summarizing the performance of all the controllers (including those in the appendix, Dan's controllers, RL with partial observability, RL with full observability).}
}

\begin{figure*}[t]
% \centering
% trim: left bottom right top
\includegraphics[width=0.315\textwidth, trim=0 0 40 0, clip]{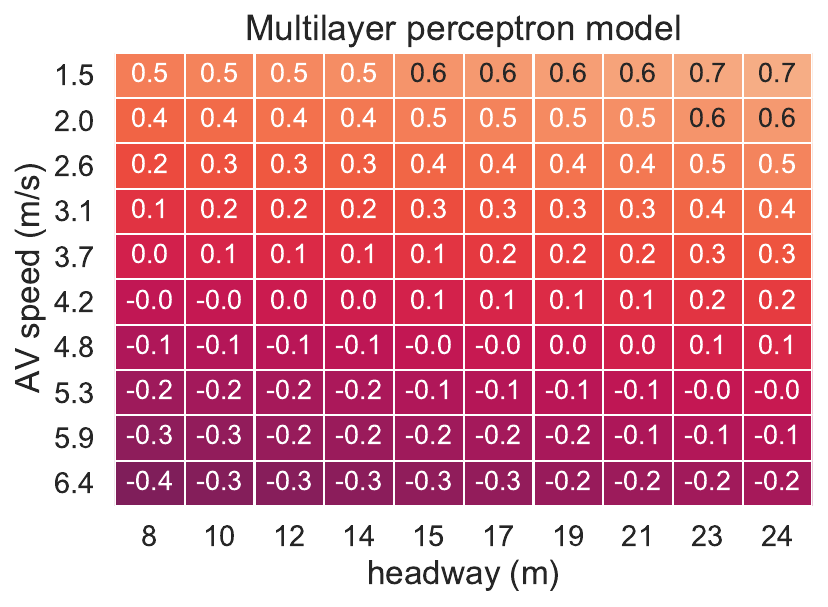}
\includegraphics[width=0.315\textwidth, trim=0 0 40 0, clip]{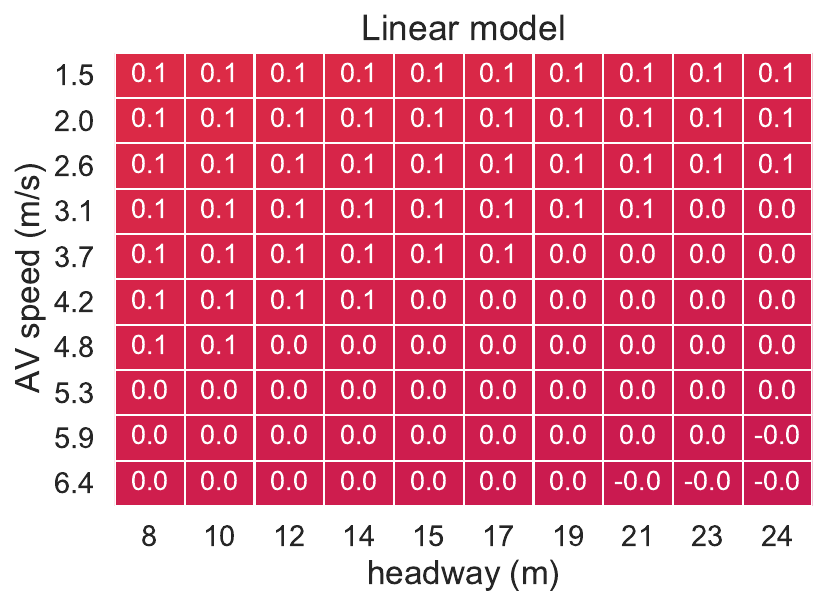}
\includegraphics[width=0.367\textwidth, trim=0 0 0 0, clip]{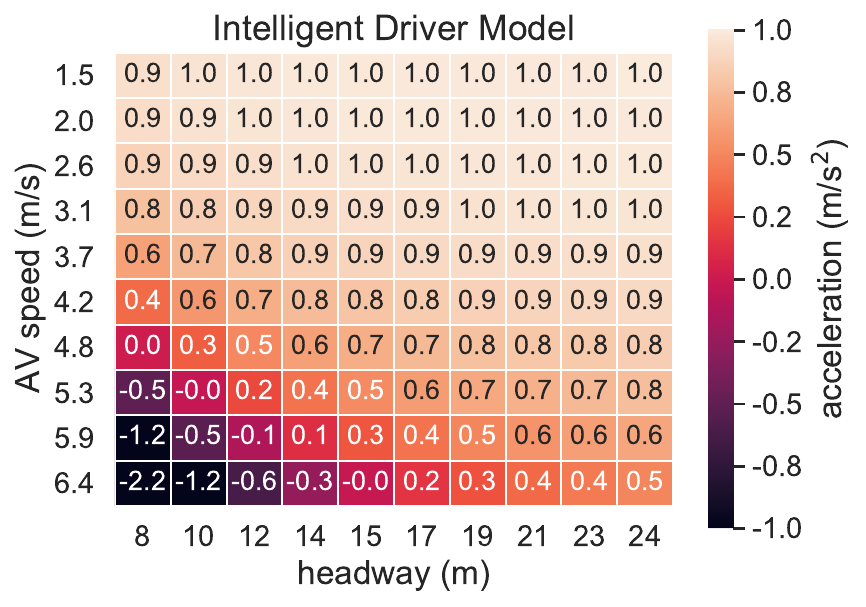}
\caption{\footnotesize \textbf{Visualization of vehicle control laws.} The heatmaps are 2-dimensional slices of the controllers (3-dimensional), and the color depicts the output (acceleration). The x-axis is a representative range of headways seen by vehicles during training. The y-axis is a representative range of AV speeds. Displayed is the slice of acceleration values of the model when the leader vehicle speed is fixed at 4.2 m/s (a typical speed for the 250 m track). The single colorbar is shared by all plots. \textbf{Left}: Learned MLP model, with failsafes disabled. \textbf{Middle}: Learned Linear model, with failsafes enabled. \textbf{Right}: IDM.}
\label{fig:heatmaps}
\end{figure*}

\newCathyTwo{
\subsubsection{Interpreting the controllers}
Yet another advantage of the partially observed setting is that the low dimensionality of the observation space lends itself to interpretation.
In Figure~\ref{fig:heatmaps}, we illustrate differences between the learned controllers and IDM.
We use a heatmap to show 2-dimensional slices of the controllers of 3-dimensional inputs, and the color of the heatmap represents the output (acceleration).
From the heatmap, we can see that the MLP controller (left subfigure) generally speeds up when it is slower than its leader and slows down when it is faster.
However, it will also increase its speed when faster if it is far away.
Notable for the MLP controller is the ``0.0'' entry in the left heatmap; it corresponds closely to the uniform flow equilibrium density. The controller can be interpreted as regulating its speed and headway such that its speed (4.2 m/s) matches the speed of its leader (4.2 m/s) at a specific density (corresponding to headway of 12 m).

On the other hand, the Linear controller is minimally reactive (middle subfigure); 
across the board of speeds and headways, the control law issues very small accelerations (and decelerations).
However, note that the Linear controller is still nonlinear due to the failsafe mechanisms built into the traffic simulator, which prevent vehicle crashes.
That is, the controller is overridden by the simulator whenever the AV's headway is too small.
Without enabling failsafe mechanisms, we found in separate experiments (not shown) that the Linear controller typically converges to a controller with frequent collisions or significantly lower performance.
Additionally, there may be further sources of nonlinearity that are introduced through the learning algorithm (e.g. observation and action clipping), and this warrants further investigation.
Our experiments indicate that a Linear model, even with nonlinearities introduced by failsafes and otherwise, may not be able to achieve the optimal velocity for the mixed autonomy circular track.
The MLP controller (left), is similarly prone to exploiting the simulator failsafes, and thus for ease of interpretation, the heatmap is displaying a controller trained without failsafes enabled.

In comparison to both learned controllers, IDM (right subfigure) is visibly more aggressive in its acceleration and deceleration. % (larger values).
Even when the vehicle is faster than its leader, it continues to accelerate until its headway is very small. This behavior, sensibly, results in stop-and-go traffic.

}

\section{Reusable modules for Mixed Autonomy}
\label{sec:flow-experiments}

\newCathy{

The previous section showed that the modules presented in Section~\ref{sec:modules} can be composed \newCathyTwo{to study open problems in traffic control} and to rigorously \newCathyTwo{evaluate RL and model-based approaches}.
\newCathyTwo{Additionally, the capacity of RL to optimally solve the idealized circular track scenario in Section~\ref{sec:mixed-autonomy-ring} provides motivation to build more complex traffic scenarios and study the mixed autonomy performance with RL.}
This section goes beyond commonly studied scenarios and demonstrates that the modules can be configured to create new scenarios with important traffic characteristics, such as multiple AVs interacting, lane changes, and intersections.
\newCathyTwo{While larger scale scenarios can also be composed, they are out of scope of this article, due to the sample efficiency limitations of current deep RL methods; this is an important direction of future work.}
\newCathyTwo{Instead,} we present several scenarios to demonstrate the richness of composing simple modules and the insights that can be derived from training controllers \newCathyTwo{therein}.
% The construction of larger networks with more vehicles or for specific traffic contexts is out of scope of this article, and are additionally limited by the computational efficiency and explainability of current reinforcement learning methods, and are important directions of research.

Notably, in contrast to \newCathyTwo{typical} model-based control approaches \newCathyTwo{to traffic control}, \newCathyTwo{RL requires} limited domain knowledge or analysis for the study of these more complex traffic control problems.
\newCathyTwo{In contrast to sophisticated mathematical analysis, we instead need to design a suitable reward function, which does require some degree of trial-and-error.
Ultimately, we selected a sensible reward function as before, focusing on system velocity and a secondary control cost term.
The control cost will vary depending on the type of control action.}

\newCathyTwo{For brevity, we describe} only the differences \newCathyTwo{relative to the modules described} in Section~\ref{case-study:mixed-autonomy-ring}.
%in order to produce a control law for a new traffic control task.
% That is, modules not mentioned are plug-and-play from Section~\ref{case-study:mixed-autonomy-ring}.
All methods are compared against a baseline of human performance \newCathyTwo{(IDM), as there are no known AV control laws for the following scenarios.}
% as there are limited existing mixed autonomy results as we explore more complex scenarios.
Results are summarized in Table~\ref{table:experiment-results-summary}.
\newCathyTwo{Based on the findings of Section~\ref{case-study:mixed-autonomy-ring},} these experiments use a memory-less diagonal Gaussian MLP control law, with hidden layers (100, 50, 25) and \texttt{tanh} non-linearity.
}

% The experiments are run on \textit{Amazon Web Services} (AWS) \textit{Elastic Compute Cloud} (EC2) instances of model \texttt{c4.2xlarge}, which have eight CPUs and 15 GB of memory.

% \todoCathy{Add a table which summarizes the results in this section (perhaps include the previous section too). Include a \% improvement column and a \% AVs column.}
% \todoAboudy{done}

%% Calculations for table
% Single lane
% (2.47-1.81)/1.81 = 36.46
% (6.02-4.2688)/4.2688 = 41.02
% (2.4707-2.545)/2.545 = -2.92
% (6.025588-6.174)/6.174 = -2.40

% Extra experiments
% (3.70-2.3996)/2.3996 = 54.19
% (4.44-2.3996)/2.3996 = 85.03
% (3.66-2.3996)/2.3996 = 52.53

\begin{table}[]
\centering
\begin{tabular}{ |l|l|l|l|l|l| } 
 \hline
 \textbf{Network} & \centertab{\#} & \centertab{\#} & \textbf{improvement} & \centertab{improvement} \\
 & \textbf{vehicles} & \textbf{AVs} & \centertab{vs human} & \textbf{vs uniform flow} \\
 \hline
 \ref{sec:mixed-autonomy-ring}  & 22  & 1   & 36.46-41.02\%  & - 2.92-2.40\%  \\ 
 \ref{sec:single-lane-multi-av}   & 22  & 3   & 54.19\% & 4.07\%   \\ 
 \ref{sec:single-lane-multi-av}   & 22  & 11  & 85.03\% & 27.07\%  \\ 
 \ref{sec:multi-lane-multi-av}    & 44  & 6   & 52.53\% & 6.09\%   \\
 \ref{sec:figure-eight-multi-av}  & 14  & 1   & 57.45\% & -- \\
 \ref{sec:figure-eight-multi-av}  & 14  & 14  & 150.49\% & -- \\
 \hline
\end{tabular}
\caption{\textbf{Mixed autonomy performance summary.} Performance improvement is given with respect to overall average velocity. Improvements over human driving (IDM) are given for all scenarios. Where possible, improvements over the uniform flow equilibrium are also given.
% are  For the \ref{sec:single-lane-multi-av} and \ref{sec:multi-lane-multi-av} experiments, the average velocity improvement (avg. speed) is calculated relative to the uniform flow equilibrium. For the \ref{sec:figure-eight-multi-av} experiment, the average velocity improvement is calculated relative to the no-autonomy setting.
}
\label{table:experiment-results-summary}
\vspace{-10pt}
\end{table}

\newCathy{
\subsection{Single-lane track with multiple autonomous vehicles} \label{sec:single-lane-multi-av}
% \todoEugene{Do you need ot define the rewards in these settings?}

In this section, a simple extension to the previous experiment shows that many variants to the same problem may be analyzed numerically, bypassing the need to adhere to strict mathematical frameworks for analysis.
The following shows additionally that even simple extensions can yield interesting and significant performance improvements.
Here we describe the experimental modules, as they differ from the previous experiment.

\smallskip \noindent \textbf{Networks:} A single-lane track with fixed length $L=230$ m. % , as displayed in Figure~\ref{fig:flow-networks} (top left).

\smallskip \noindent \textbf{Observer:} All vehicle positions and velocities\newCathyTwo{, that is $o = s = (x_1, v_1, x_2, v_2,\dots, x_{n}, v_{n})$.}

\smallskip \noindent \textbf{Control laws:} Three to eleven of the actors are dictated by a single \newCathyTwo{(centralized)} learned control law\newCathyTwo{; that is, $\mathcal{A} = \mathbb{R}^m$ where $m \in \{3, \dots, 11\}$ and $\pi_\theta: \mathcal{O} \times \mathcal{A} \to \Delta^{m}$}. \newCathyTwo{The remaining actors are modeled as human drivers according to IDM.}

\smallskip \noindent \newCathyTwo{\textbf{Metrics:} The reward function is a weighted combination of the average velocity of all vehicles and a control cost.
\begin{equation}
r(s,a) = \frac{1}{n} \sum_{i} v_i - \alpha \frac{1}{m} \sum_{j \in [m]} |a_j|
\end{equation}
where $\alpha = 0.1$}.

\smallskip \noindent \textbf{Result:}} A string of consecutive AVs learns to proceed with a smaller headway than the human driver models (platooning), resulting in greater roadway utilization, thereby surpassing the upper bound from Section~\ref{sec:mixed-autonomy-ring},
% permitting a higher velocity for the overall system, 
as can be seen in Figure~\ref{fig:platooning-vel}.

% \begin{figure}[th]
% \centering
% \includegraphics[width=0.4\textwidth]{figures/platooning-headway.png}
%   \caption{Headway profile in the presence of 11 autonomous vehicles. The autonomous vehicles in the network possess an average headway of 4.5m amongst themselves, far below the average human headway of 6.5m.}
% \label{fig:platooning-headway}
% \end{figure}

\begin{figure}[th]
\centering
\pgfplotstableread[col sep = comma]{data/platooning_vel.csv}\mydata
\begin{tikzpicture}
	\begin{axis}[
		legend style={nodes={scale=0.75, transform shape}},
		legend pos=south east,
		height=6cm, 
		width=9cm, 
		grid=major,
		ylabel=velocity (m/s), 
		ylabel near ticks,
        xlabel= time (s), 
		xticklabel style={
  			/pgf/number format/precision=3,
  			/pgf/number format/fixed
  		},
  		label style={font=\small},
  		tick label style={font=\small}
    ]

	\addplot[black, dashed] coordinates {(0,3.45406618) (150,3.45406618)};
    \addlegendentry{uniform flow equilibrium}

    \addplot[red] table[x index = {0}, y index = {1}]{\mydata};
    \addlegendentry{3 autonomous vehicles}[fontsize=1cm]

    \addplot[black!30!blue] table[x index = {0}, y index = {2}]{\mydata};
    \addlegendentry{11 autonomous vehicles}
 
	\end{axis}
\end{tikzpicture}
  \caption{Velocity profile for single-lane \newCathy{circular track} with multiple AVs. With additional AVs, the average velocity exceeds the uniform flow equilibrium velocity, and continues to increases as the number of AVs increase. At three AVs, the average velocity settles at 3.70 m/s; at 11 AVs, the average velocity settles at 4.44 m/s.}
\label{fig:platooning-vel}
\vspace{-10pt}
\end{figure}

\newCathy{

\subsection{Multi-lane track with multiple autonomous vehicles}
\label{sec:multi-lane-multi-av}

Multi-lane settings are challenging to study from a model-based control-theoretic perspective due to the discontinuity in model dynamics from lane changes, as well as due to the complexity of mathematically modeling lane changes accurately.
The experimental modules are described here:

\smallskip \noindent \textbf{Networks:} The network is a two-lane circular track of \newCathyTwo{$L=230$ m}, as displayed in Figure~\ref{fig:flow-networks} (top center).

\smallskip \noindent \textbf{Actors:} There are \newCathyTwo{$n=44$ vehicles}.

\smallskip \noindent \textbf{Observer:} All vehicle positions, velocities, and lanes\newCathyTwo{, that is $o = s = (x_1, v_1, l_1, x_2, v_2, l_2, \dots, x_{n}, v_{n}, l_{n})$}.

\smallskip \noindent \textbf{Control laws:} Six of the actors are dictated by a single learned control law for both acceleration and lane changes (continuous action representation)\newCathyTwo{; that is, $\mathcal{A} = \mathbb{R}^{2m}$ for $m=6$}. The rest are dictated by IDM for longitudinal control and the lane changing model of SUMO for lateral control.

\smallskip \noindent \textbf{Initialization:} The vehicles are evenly spaced in the track, using both lanes, and six AVs are placed together in sequence, all in the outer lane.

\smallskip \noindent \newCathyTwo{\textbf{Metrics:} The reward function is a weighted combination of the average velocity of all vehicles and the a control cost.
\begin{equation}
r(s,a) = \frac{1}{n} \sum_{i} v_i - \alpha \frac{1}{2m} \sum_{j \in [2m]} |a_{j}|
\end{equation}
where $\alpha = 0.1$. Note that both excessive accelerations and lane changes are penalized}.

\smallskip \noindent \textbf{Result:}
The learned control law yields AVs balancing across the two lanes, with three in each lane, and avoiding stop-and-go waves in both lanes.
The resulting average velocity is 3.66 m/s, an improvement over the 3.45 m/s uniform flow equilibrium velocity.
Even though the control inputs are a mix of inherently continuous and discrete signals (acceleration and lane change, respectively), a well-performing control law was learned using only a continuous representation, which is a testament to the flexibility of the approach.

}

\eat{In a multi-lane setting, in addition to platooning together, the AVs learn to evenly distribute themselves among the lanes in order to ensure that all vehicles in the network benefit from the same level of platooning.}

\eat{
\todoAboudy{TODO: single lane platooning with multiple penetrations, multi lane platooning, multi-agent single lane platooning}
\todoAboudy{Plots: single lane platooning velocities for different percentages of autonomous penetration, comparison of performance of platooning in single lane, multi-lane, and multi-agent settings, 4 space-time diagrams: all human, stabilization result, platooning, all AVs}
}

\eat{
\todoCathy{Give performance curve: velocity vs number of percent of autonomous vehicles.}
\todoCathy{Give 4 space-time diagrams: all human, stabilization result, platooning, all AVs.}
}

\subsection{Intersection with mixed and full autonomy} \label{sec:figure-eight-multi-av}
\eat{
\todoCathy{Give performance curve: velocity vs number of percent of autonomous vehicles.}
\todoCathy{Give 3 space-time diagrams: all human, 1 AV, all AV.}}

We now consider a simplified intersection scenario, which demonstrates the ease of considering different network topologies and traffic rules, such as right-of-way rules at intersections.
The authors are not aware of any model-based mixed autonomy results for intersections, with which to compare.
In the absence of autonomous vehicles, human drivers queue at the intersection, leading to significant delays; this serves as our baseline.
% The proposed methodology may \newCathyTwo{of course} also be used to study full autonomy settings, in addition to mixed autonomy settings.
We now describe the traffic control scenario:

\smallskip \noindent \textbf{Networks:} The network resembles a figure eight, as displayed in Figure~\ref{fig:flow-networks} (top right), with an intersection in the middle, two circular tracks of radius 30 m, and total length of 402 m.

\smallskip \noindent \textbf{Actors:} There are \newCathyTwo{$n=14$ vehicles}.

\smallskip \noindent \textbf{Observer:} All vehicle positions and velocities\newCathyTwo{, that is $o = s = (x_1, v_1, x_2, v_2,\dots, x_{n}, v_{n})$}.

\smallskip \noindent \textbf{Control laws:} One or all of the actors are dictated by a single learned control law for acceleration control inputs\newCathyTwo{; that is, $\mathcal{A} = \mathbb{R}$ or $\mathcal{A} = \mathbb{R}^m$}. The rest are dictated by IDM for longitudinal control and intersection rules from SUMO.

\smallskip \noindent \textbf{Dynamics:} There is no traffic light at the intersection; instead vehicles crossing the intersection follow SUMO's right-of-way model to enforce traffic rules and to prevent crashes.

\smallskip \noindent \newCathyTwo{\textbf{Metrics:} The reward function is a weighted combination of the average velocity of all vehicles and the a control cost.
\begin{equation}
r(s,a) = \frac{1}{n} \sum_{i} v_i - \alpha \frac{1}{m} \sum_{j \in [m]} |a_j|
\end{equation}
where $\alpha = 0.1$}.

\smallskip \noindent \textbf{Result:} With \newCathyTwo{only} one autonomous vehicle, the learned control law results in the vehicles moving on average 1.5 times faster. The full autonomy setting exhibits an improvement of almost three times faster. Figure~\ref{fig:figure8-vel} shows the average velocity of vehicles in the network for different levels of autonomy.

\begin{figure}[th]
\centering
\pgfplotstableread[col sep = comma]{data/fig8_vel.csv}\mydata
\begin{tikzpicture}
	\begin{axis}[
		legend style={nodes={scale=0.75, transform shape}},
		legend pos=south east,
		height=6cm, 
		width=9cm, 
		grid=major,
		ylabel=velocity (m/s), 
		ylabel near ticks,
        xlabel= time (s), 
		xticklabel style={
  			/pgf/number format/precision=3,
  			/pgf/number format/fixed
  		},
  		label style={font=\small},
  		tick label style={font=\small}
    ]

    \addplot[red] table[x index = {0}, y index = {1}]{\mydata};
    \addlegendentry{0 autonomous vehicles}[fontsize=1cm]

    \addplot[black!30!green] table[x index = {0}, y index = {2}]{\mydata};
    \addlegendentry{1 autonomous vehicles}

    \addplot[blue] table[x index = {0}, y index = {3}]{\mydata};
    \addlegendentry{14 autonomous vehicles}
 
	\end{axis}
\end{tikzpicture}
  \caption{Velocity profile for intersection scenarios, with different fractions of AVs. The performance improves as the number of AVs increases; even one autonomous vehicle results in a velocity improvement of 1.5x, and full autonomy almost triples the average velocity from 5 m/s (no autonomy) to 14 m/s.}
\label{fig:figure8-vel}
% \vspace{-10pt}
\end{figure}

Further study is needed to understand, interpret, and analyze the above learned behaviors and control laws, and thereby take steps towards a real-world deployment and policy analysis.
Some preliminary investigations can be found in~\cite{Wu2017d, kreidieh2018dissipating}.

\section{Conclusion}
\label{sec:conclusion}

The complex integration of autonomy into existing systems introduces new technical challenges beyond studying autonomy in isolation or in full.
This article aims to make progress on these upcoming challenges by studying how modern deep RL can be leveraged to gain insights into complex mixed autonomy systems.
In particular, we focus on the integration of autonomous vehicles into urban systems, as we expect these to be among the first robotic systems to enter and affect existing societal systems.
The article introduces Flow, a modular learning framework which eases the composition of modules, to enable learning control laws for AVs in complex traffic scenarios. % involving nonlinear vehicle dynamics and arbitrary network configurations.
Several experiments yielded controllers which far exceeded state-of-the-art performance (in fact, achieving near-optimal performance) and demonstrated the generality of the methodology for disparate traffic dynamics.
Since an early version of this manuscript was made available online in 2017~\cite{wu2017flow}, several works have employed this framework to achieve results in the directions of the discovery of emergent behaviors in traffic, transfer learning, bottleneck control, the design of traffic control benchmarks, and sim2real~\cite{Wu2017d, kreidieh2018dissipating, vinitsky2018lagrangian, vinitsky2018benchmarks, jang2019simulation}.

Open directions of research include the study of mixed autonomy in larger and more complex traffic networks; studying different, more realistic, and regional objective functions; studying delayed vehicle models; devising new RL techniques for large-scale networked systems; studying scenarios with variable numbers of actors; incorporating advances in safe RL; and incorporating multi-agent RL for the study of variable numbers of autonomous vehicles. 
Another open direction is to study fundamental limitations of this methodology; in particular, 1) establishing when guarantees of global convergence, stability, robustness, and safety of classical approaches can not be achieved with deep RL, and 2) quantifying the effects of simulation model error or misspecification on training outcomes.
We additionally plan to extend Flow with modules suitable for the study of other forms of automation in traffic, such as problems concerning traffic lights, road directionality, signage, roadway pricing, and infrastructure communication.
Another interesting direction is whether an analogous design of reusable modules may be used for other robotics and transportation-related scenarios, such as for motion planning, navigation, ridesharing, and land use planning.

\section*{Acknowledgments}
\begin{footnotesize}
The authors would like to thank Professor Alexander Skabardonis for insightful discussions about vehicle dynamics; Leah Dickstein and Nathan Mandi for early prototyping; Nishant Kheterpal, Kathy Jang, Saleh Albeaik and Ananth Kuchibhotla for helping to build out features; Jakob Erdmann for SUMO support; Rocky Duan and Alex Lee for rllab support, used in early experiments; Philipp Moritz, Eric Liang, and Richard Liaw for Ray and RLlib support; Zhongxia Yan for discussions and training framework support; and Joseph Wu for detailed comments on the manuscript.
The team is grateful to Professor Dan Work for technical conversations about the circular track experiment, and to the inspirational work of the Piccoli-Seibold-Sprinkle-Work team.
The authors would additionally like to thank the anonymous reviewers of early versions of the manuscript for their insightful and constructive comments.
\end{footnotesize}

\appendices

\section{Networks}
\label{sec:networks}
Flow supports learning policies on arbitrary (user-defined) networks.
In this section, we include a basic set of networks, which we have designed or adapted from the literature to capture important traffic phenomena.
These include \textit{closed networks}, such as single and multi-lane circular tracks, figure-eight networks, and loops with merge, as well as \textit{open networks}, such as intersections, merge networks and highway networks.
In contrast to closed networks, open networks require pre-defined in-flows of vehicles into the traffic system.
See Figure~\ref{fig:flow-networks} for various example networks supported by Flow.
In each of these networks, Flow can be used to study the design or learning of controllers which optimize the system-level velocity or other objectives, in the presence of different types of vehicles, model noise, etc.

\eat{\todoAboudy{Can we add citations? I think mentioning how much this is studied ties in well with the multi-lane ring section}}

\smallskip \noindent \textbf{Single-lane circular tracks:} This network consists of a circular lane with a specified length, inspired by the 230m track studied by Sugiyama et al.~\cite{Sugiyama2008}. This canonical benchmark has been extensively studied. % and serves as a canonical benchmark for traffic control. % n experimental and numerical baseline for benchmarking.

\smallskip \noindent \textbf{Multi-lane circular tracks:}
Multi-lane circular tracks are a natural extension to the single-lane track. The inclusion of lane-changing behavior in this setting makes studying such problems exceedingly difficult from an analytical perspective, thereby constraining most classical control techniques to the single-lane case. Many multi-lane models forgo longitudinal dynamics in order to encourage tractable analysis~\cite{Michalopoulos1984, Klar1998, Sasoh2002, Daganzo2002a}. Recent strides have been made in developing simple stochastic models that retain longitudinal dynamics while capturing lane-changing dynamics in a single lane setting~\cite{Wu2017b}. Modern machine learning methods, however, do not require a simplification of the dynamics for the problem to become tractable, as explored in Section~\ref{sec:multi-lane-multi-av}.

\smallskip \noindent \textbf{Figure-eight network:}
The figure-eight network is a simple closed network with an intersection. Two circular tracks, placed at opposite ends of the network, are connected by two perpendicular road that cross at an intersection. Vehicles that try to cross this intersection from opposite ends are constrained by SUMO's right-of-way model to prevent crashes.

\smallskip \noindent \textbf{Loops with merge network:}
This network permits the study of merging behavior within a closed network.
This network consists of two circular tracks which are connected together.
Vehicles in the smaller track stay within this track, while vehicles in the larger track try to merge into the smaller track and then back out to the larger track.
This typically results in congestion at the merge points.

\smallskip \noindent \textbf{Intersections:} 
This network permits the study of intersection management in an open network.
Vehicles arrive in the control zone of the intersection according to a Poisson distribution.
At the control zone, the system speeds or slows down vehicles to either maximize average velocity or minimize experienced delay.
This building block can be used to build a general schema for arbitrary maps such as the one shown in Figure~\ref{fig:flow-networks} (bottom right).

\section{Intelligent Driver Model}
\label{sec:idm}
The \textit{Intelligent Driver Model} (IDM) is a car following model capable of accurately representing realistic driver behavior~\cite{Treiber2000} and reproducing traffic waves, and is commonly used in the transportation research community.
We employ IDM in the numerical experiments of this article, and therefore analyze this specific model to compute the theoretical performance bounds of the overall traffic system.
The acceleration for a vehicle modeled by IDM is defined by: % its bumper-to-bumper headway $h$ (distance to preceding vehicle), velocity $v$, and relative velocity $\dot h$, via the following equation:
\begin{equation} \label{eq:idm}
a_{\text{IDM}} = \frac{dv}{dt} = a \bigg[ 1 - \bigg( \frac{v}{v_0} \bigg)^\delta - \bigg( \frac{H(v,\dot h)}{h} \bigg)^2 \bigg],
\end{equation}
where $H(\cdot)$ is the desired headway of the vehicle, denoted by:
\begin{equation} \label{eq:s_star}
H(v,\dot h) = s_0 + \max \bigg( 0, v T + \frac{v \dot h}{2 \sqrt{ab}} \bigg),
\end{equation}
where $h_0, v_0, T, \delta, a, b$ are given parameters.
Table~\ref{table:controller-params} % describes the parameters of the model and 
provides typical parameters for highway driving~\cite{treiber2013traffic}.

\begin{table}[h!]
\scriptsize
\centering
\begin{tabular}{ |l|l|l|l|l|l|l|l|l|l|l|l|l|l|l|l|l|l|l|l|l|l|l|l|l|l| } 
 \hline
                     & \multicolumn{7}{c|}{Intelligent Driver Model (IDM)} \\
 \hline
 \textbf{Parameter} & $v_0$ & $T$ & $a$ & $b$ & $\delta$ & $h_0$ & noise \\
 \hline
 \textbf{Value}     & 30 m/s & 1 s & 1 m/s$^2$ & 1.5 m/s$^2$ & 4 & 2 m & $\mathcal{N}(0,0.2)$ \\
 \hline
\end{tabular}\\[10pt]
\caption{\newCathy{Parameters for car-following control law}}
\label{table:controller-params}
\end{table}

\section{Model-based longitudinal control laws}
\label{sec:controllers}

\newCathy{
In this section, we detail the two state-of-the-art control laws for the mixed autonomy circular track, against which we benchmark our learned policies generated using Flow.}

\setcounter{subsubsection}{0}

\subsubsection{FollowerStopper}
\label{sec:stopper-follower}

\begin{table*}[h!]
\centering

\begin{tabular}{ |l|l|l|l|l|l|l|l|l|l| } 
 \hline
                     & \multicolumn{4}{c|}{PI with Saturation} \\
 \hline
 \textbf{Parameters} & $\delta$ & $g_l$ & $g_u$ & $v^\text{catch}$\\
 \hline
 \textbf{Values}     & 2 m & 7 m & 30 m & 1 m/s \\
 \hline
\end{tabular}\\[10pt]

\begin{tabular}{ |l|l|l|l|l|l|l|l|l|l|l|l|l|l|l|l|l|l|l|l|l|l|l|l|l|l| } 
 \hline
                     & \multicolumn{11}{c|}{FollowerStopper} \\
 \hline
 \textbf{Parameters} & $\Delta x_1^0$ & $\Delta x_2^0$ & $\Delta x_3^0$ & $d_1$ & $d_2$ & $d_3$ & $\Delta v$ & $\Delta x_1$ & $\Delta x_2$ & $\Delta x_3$ & $U$
 \\
 \hline
 \textbf{Values}     & 4.5 m & 5.25 m & 6.0 m & 1.5 m/s$^2$ & 1.0 m/s$^2$ & 0.5 m/s$^2$ & -3 m/s & 7.5 m/s & 9.75 m/s & 15 m/s & 4.15 m/s  \\
 \hline
\end{tabular}

\caption{\newCathy{Parameters for model-based controllers}}
\label{table:controller-params-model-based}
\end{table*}

Recent work by~\cite{stern2018dissipation} presented two control models that may be used by autonomous vehicles to attenuate the emergence of stop-and-go waves in a traffic network. The first of these models is the \textit{FollowerStopper}. This model commands the AVs to maintain a desired velocity $U$, while ensuring that the vehicle does not crash into the vehicle behind it. Following this model, the command velocity $v^\text{cmd}$ of the autonomous vehicle is:
\begin{equation}
v^{\text{cmd}} = 
\begin{cases}
0 & \text{if } \Delta x \leq \Delta x_1 \\
v \frac{\Delta x - \Delta x_1}{\Delta x_2 - \Delta x_1} & \text{if } \Delta x_1 < \Delta x \leq \Delta x_2 \\
v + (U - v) \frac{\Delta x - \Delta x_2}{\Delta x_3 - \Delta x_2} & \text{if } \Delta x_2 < \Delta x \leq \Delta x_3 \\
U & \text{if } \Delta x_3 < \Delta x
\end{cases}
\end{equation}

where $v = \min(\max (v^\text{lead}, 0), U)$, $v^\text{lead}$ is the speed of the leading vehicles, $\Delta x$ is the headway of the autonomous vehicle, subject to boundaries defined as:
\begin{equation}
\Delta x_k = \Delta x_k^0 + \frac{1}{2 d_k} (\Delta v_-)^2, \ \ k=1,2,3
\end{equation}

The parameters of this model can be found in~\cite{stern2018dissipation} and are also provided in Table~\ref{table:controller-params-model-based}.

\subsubsection{PI with Saturation}
\label{sec:pi-controller}
In addition to the \textit{FollowerStopper} control law,~\cite{stern2018dissipation}~presents a model called the \textit{PI with Saturation control law} that attempts to estimate the average equilibrium velocity $U$ for vehicles on the network, and then drives at that speed. This average is computed as a temporal average from its own history: $U = \frac{1}{m} \sum_{j=1}^m v_j^{AV}$. The target velocity at any given time is then defined as:
\begin{equation}
v^{\text{target}} = U + v^{\text{catch}} \times \min \left(\max \left(\frac{\Delta x - g_l}{g_u - g_l},0 \right), 1 \right)
\end{equation}

Finally, the command velocity for the vehicle at time $j+1$, which also ensures that the vehicle does not crash, is:
\begin{equation}
v_{j+1}^{\text{cmd}} = \beta_j (\alpha_j v_j^{\text{target}} + (1-\alpha_j)v_j^{\text{lead}}) + (1-\beta_j)v_j^{\text{cmd}}
\end{equation}

The values for all parameters in the model can be found in~\cite{stern2018dissipation} \newCathy{and are also provided in Table~\ref{table:controller-params-model-based}}.

\bibliographystyle{IEEEtran}
% {\footnotesize
\bibliography{mybib}
% }

\vspace{-10mm}

\begin{IEEEbiography}[{\includegraphics[width=1in,height=1.25in,clip,keepaspectratio]{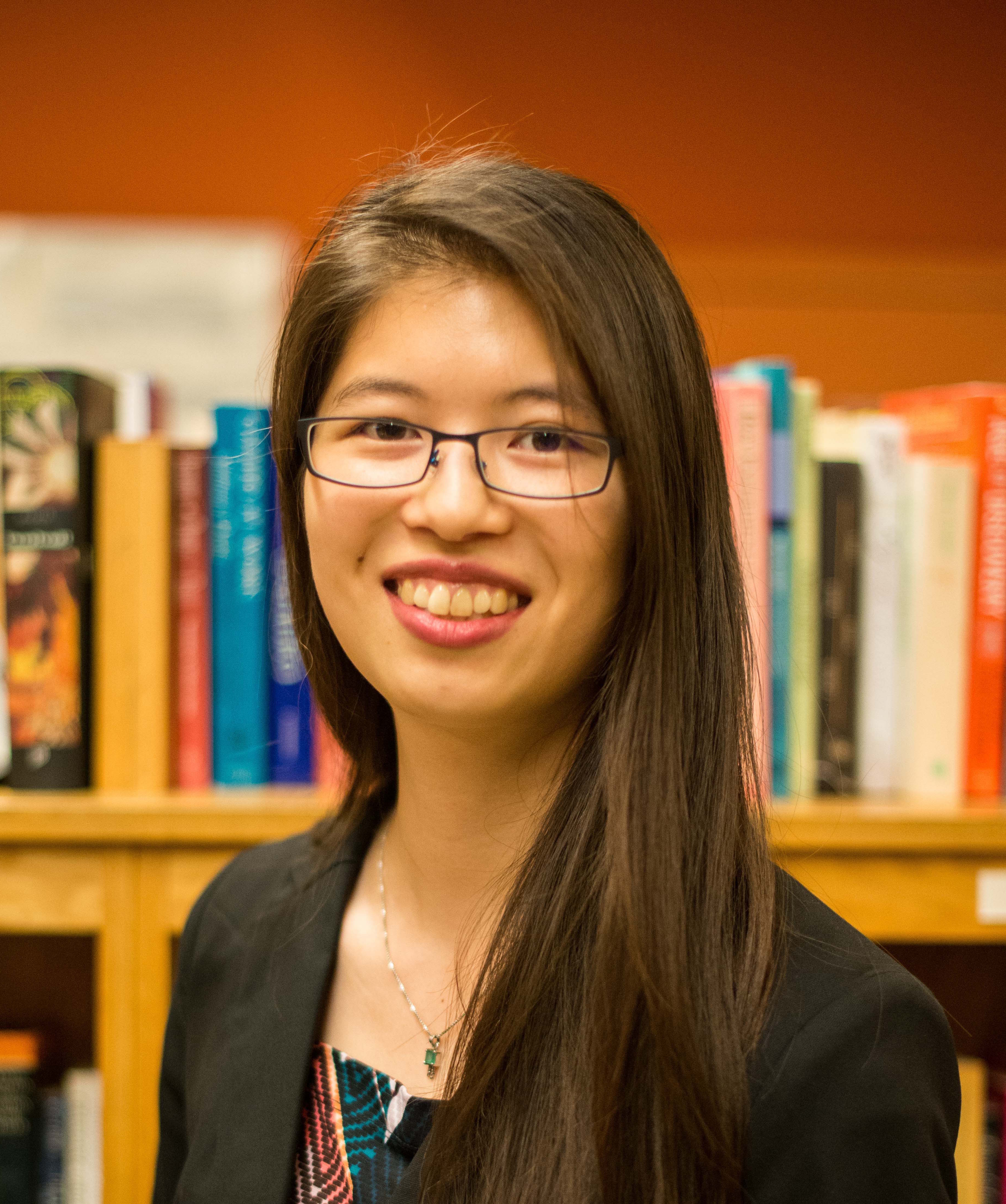}}]{Cathy Wu}
received the B.S. and M.Eng. degrees from the Massachusetts Institute of Technology (MIT), Cambridge, MA, USA, in 2012 and 2013, respectively, and the Ph.D. degree from the University of California Berkeley, Berkeley, CA, USA, in 2018, all in electrical engineering and computer sciences.

She was a Postdoc with the Microsoft Research AI. She is an Assistant Professor with MIT in LIDS, CEE, and IDSS. She studies the technical challenges surrounding the integration of autonomy into societal systems. Her research interests include machine learning and mobility.

Prof. Wu was a recipient of several awards, including the 2019 IEEE Intelligent Transportation Systems Society (ITSC) Best Ph.D. Dissertation Award, 2018 Milton Pikarsky Memorial Dissertation Award, and the 2016 IEEE ITSC Best Paper Award, and has appeared in the press, including Wired and Science.
\end{IEEEbiography}

\vspace{-10mm}

\begin{IEEEbiography}[{\includegraphics[width=1in,height=1.25in,clip,keepaspectratio]{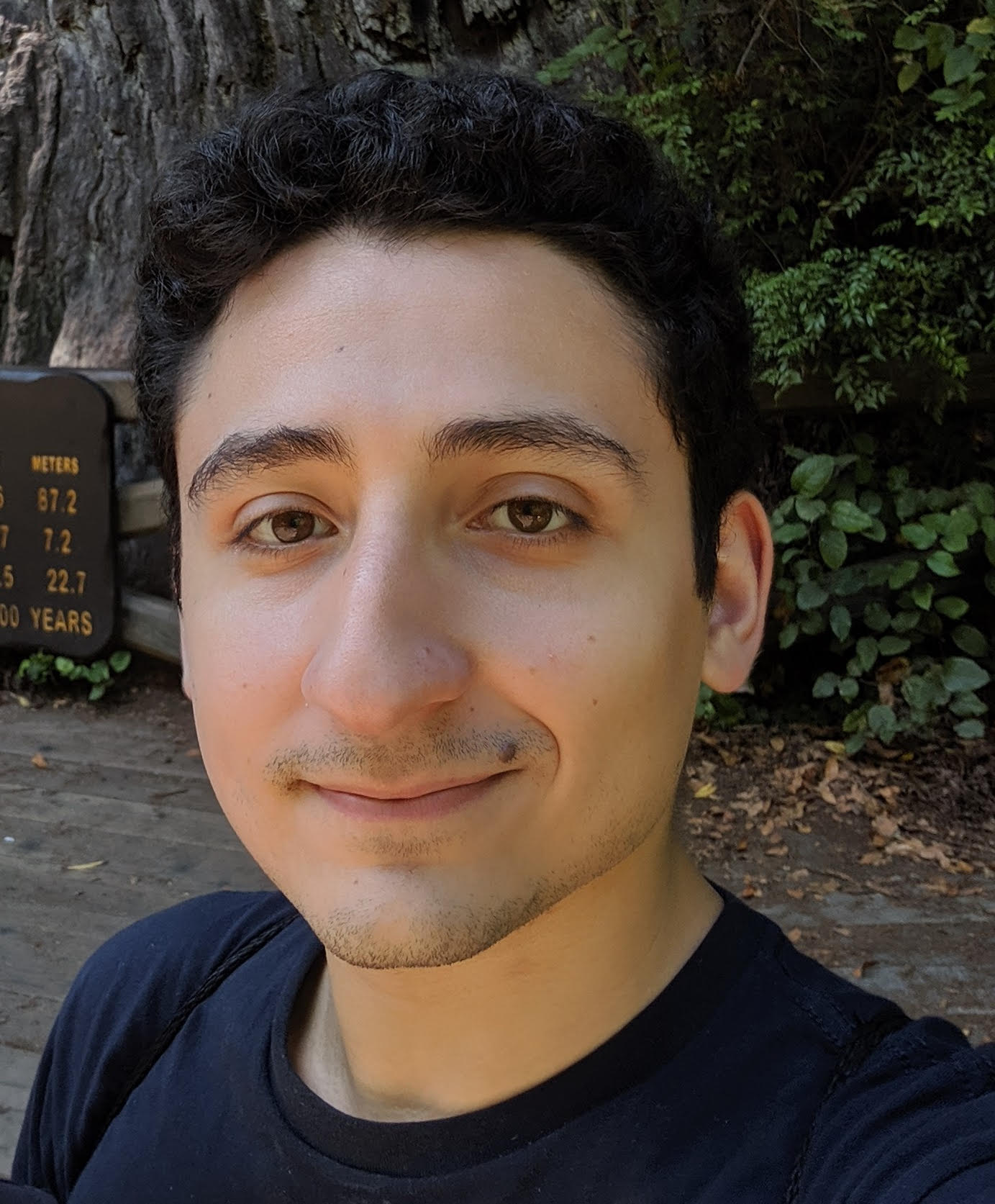}}]{Abdul Rahman Kreidieh}
received the B.S. degree in mechanical engineering from the American University at Beirut, Beirut, Lebanon, in 2016, and the M.S. degree in civil and environmental engineering in 2017 from the University of California, Berkeley, Berkeley, CA, USA, where he is currently working toward the Ph.D. degree in civil and environmental engineering.

His primary focus is on designing models and algorithms that scale the performance of existing machine learning systems to large-scale traffic control problems. His research interests include the intersection of machine learning and traffic control through mixed autonomy systems.
\end{IEEEbiography}

\vspace{-10mm}

\begin{IEEEbiography}[{\includegraphics[width=1in,height=1.25in,clip,keepaspectratio]{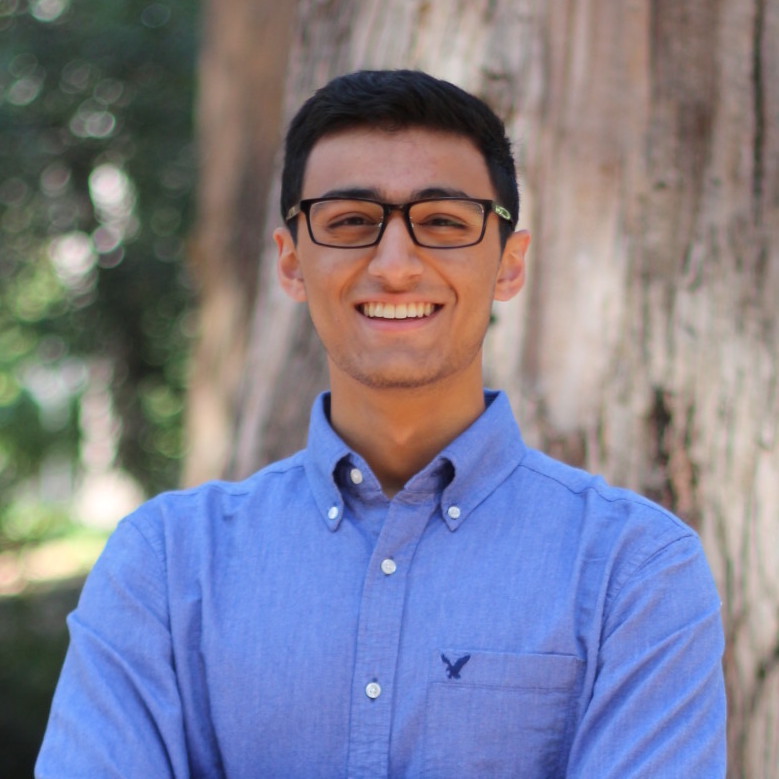}}]{Kanaad Parvate}
received the B.S. and M.S. degrees in electrical engineering and computer science from the University of California, Berkeley, Berkeley, CA, USA, in 2019.

He is currently an Engineer with Waymo on the Behavior Prediction team.
\end{IEEEbiography}

\vspace{-10mm}

\begin{IEEEbiography}[{\includegraphics[width=1in,height=1.25in,clip,keepaspectratio]{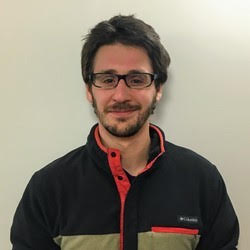}}]{Eugene Vinitsky}
received the B.S. degree in physics from the California Institute of Technology, Pasadena, CA, USA, in 2014, the M.A. degree in physics from the University of California, Santa Barbara, Santa Barbara, CA, in 2015. He is currently working toward the Ph.D. degree in controls engineering with the Mobile Sensing Laboratory, University of California, Berkeley, Berkeley, CA.

His current research interests include multiagent reinforcement learning, cooperative autonomous vehicles, and robust control.
Mr. Vinitsky is the recipient of a National Science Foundation Graduate Research Fellowship.
\end{IEEEbiography}

\vspace{-10mm}

\begin{IEEEbiography}[{\includegraphics[width=1in,height=1.25in,clip,keepaspectratio]{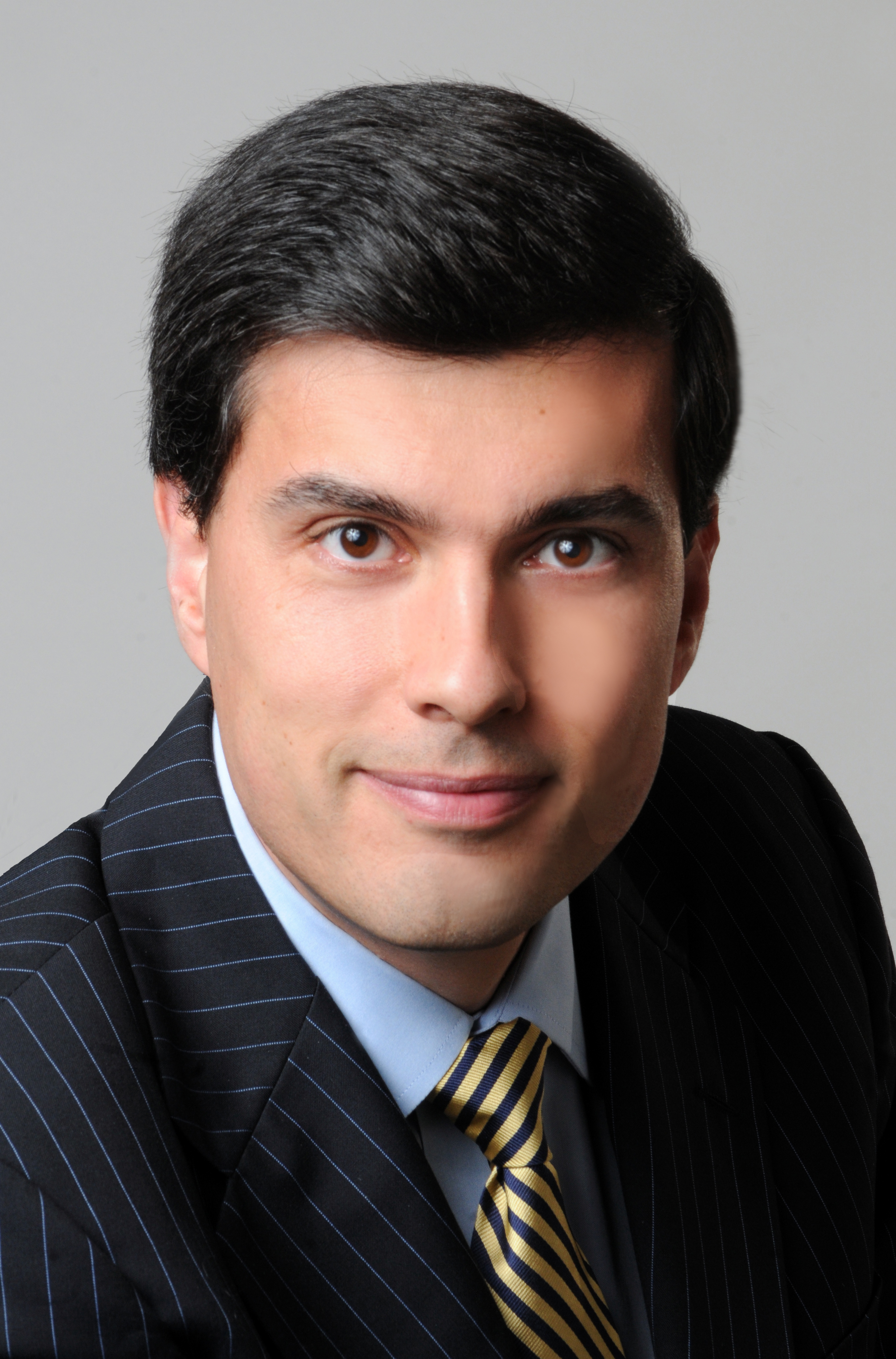}}]{Alexandre M. Bayen}
received the Engineering degree in applied mathematics from the Ecole Polytechnique, Palaiseau, France, in 1998, and the M.S. and Ph.D. degrees in aeronautics and astronautics from Stanford University, Stanford, CA, USA, in 1999 and 2004, respectively.

He is the Liao-Cho Professor of Engineering with University of California, Berkeley, Berkeley, CA, USA. He is currently a Professor of Electrical Engineering and Computer Science, and Civil and Environmental Engineering. He is currently the Director
of the Institute of Transportation Studies. He is also a Faculty Scientist in Mechanical Engineering, Lawrence Berkeley National Laboratory.
\end{IEEEbiography}

\end{document}